\documentclass{article}

\usepackage{microtype}
\usepackage{graphicx}
\usepackage{subcaption}
\usepackage{booktabs} 

\usepackage{hyperref}

\usepackage{amsfonts,amssymb,amsbsy,amsmath}
\usepackage[inline]{enumitem}
\usepackage{physics}
\usepackage{cancel}        
\usepackage{listings}      

\usepackage{soul}          
\usepackage{float}         
\usepackage{graphicx}      
\usepackage{caption}
\usepackage{subcaption}

\usepackage{xcolor}

\usepackage{hyperref} 

\usepackage{amsthm}
\newcommand{\eqn}[1]{ 
    \begin{equation}
    \begin{aligned}
        #1
    \end{aligned}
    \end{equation}
}

\renewcommand{\d}{\mathrm{d}}                                  

\renewcommand{\L}{\mathcal{L}}                                 
\renewcommand{\O}{\mathcal{O}}                                 
\renewcommand{\ul}[1]{\underline{#1}}

\usepackage{amsmath,amsfonts,bm}

\def\eqref#1{equation~\ref{#1}}

\def\1{\bm{1}}

\DeclareMathAlphabet{\mathsfit}{\encodingdefault}{\sfdefault}{m}{sl}
\SetMathAlphabet{\mathsfit}{bold}{\encodingdefault}{\sfdefault}{bx}{n}

\newcommand{\R}{\mathbb{R}}

\newcommand{\softmax}{\mathrm{softmax}}

\newcommand{\cmark}{\textcolor{green!70!black}{\checkmark}}
\newcommand{\xmark}{\textcolor{red}{\ding{55}}}
\usepackage{pifont} 

\usepackage{caption}
\usepackage{multirow}
\usepackage{colortbl}
\usepackage{makecell}

\usepackage{algorithm}
\usepackage{algorithmic}
\usepackage{wrapfig}

\usepackage{threeparttable} 

\usepackage{listings}

\definecolor{codegreen}{rgb}{0,0.6,0}
\definecolor{codegray}{rgb}{0.5,0.5,0.5}
\definecolor{codepurple}{rgb}{0.58,0,0.82}
\definecolor{backcolour}{rgb}{0.95,0.95,0.95}

\lstdefinestyle{mystyle}{
    backgroundcolor=\color{backcolour},   
    commentstyle=\color{codegreen},
    keywordstyle=\color{magenta},
    stringstyle=\color{codepurple},
    basicstyle=\ttfamily\footnotesize,
    breakatwhitespace=false,         
    breaklines=true,                 
    captionpos=b,                    
    keepspaces=true,                 
    showspaces=false,                
    showstringspaces=false,
    showtabs=false,                  
    tabsize=2
}
\newcommand{\cuttext}[1]{}

\usepackage[preprint]{icml2026}

\usepackage{amsmath}
\usepackage{amssymb}
\usepackage{mathtools}
\usepackage{amsthm}

\usepackage[capitalize,noabbrev]{cleveref}

\theoremstyle{plain}

\theoremstyle{definition}

\theoremstyle{remark}

\usepackage[textsize=tiny]{todonotes}

\icmltitlerunning{{FLARE}: {F}ast {L}ow-{r}ank {A}ttention {R}outing {E}ngine}

\begin{document}

\twocolumn[
  \icmltitle{{FLARE}: {F}ast {L}ow-{r}ank {A}ttention {R}outing {E}ngine}



  \icmlsetsymbol{equal}{*}

  \begin{icmlauthorlist}
    \icmlauthor{Vedant Puri}{equal,yyy}
    \icmlauthor{Aditya Joglekar}{equal,yyy}
    \icmlauthor{Sri Datta Ganesh Bandreddi}{yyy}
    \icmlauthor{Kevin Ferguson}{yyy}
    \icmlauthor{Yu-hsuan Chen}{yyy}
    \icmlauthor{Yongjie Jessica Zhang}{yyy}
    \icmlauthor{Levent Burak Kara}{yyy}
  \end{icmlauthorlist}

  \icmlaffiliation{yyy}{Department of Mechanical Engineering, Carnegie Mellon University, Pittsburgh, Pennsylvania, USA}

\icmlcorrespondingauthor{Vedant Puri}{vedantpu@andrew.cmu.edu}

  \icmlkeywords{Machine Learning, ICML}

  \vskip 0.3in
]



\printAffiliationsAndNotice{}  

\begin{abstract}
The quadratic complexity of self-attention limits the scalability of transformers on long sequences.
We introduce \emph{Fast Low-rank Attention Routing Engine (FLARE)}, a token-mixing operator that realizes low-rank attention by routing information through a small set of latent tokens.
Each layer induces an input-input token mixing matrix of rank at most $M$ via a minimal encode-decode factorization implemented using only two standard scaled dot-product attention (SDPA) calls.
Because the dominant $\mathcal{O}(NM)$ computation is expressed purely in terms of standard SDPA, FLARE is compatible with fused attention kernels and avoids materializing $M\times N$ projection matrices.
FLARE further assigns disjoint latent slices to each attention head, yielding a mixture of head-specific low-rank pathways.
Empirically, FLARE scales to \emph{one-million-point} unstructured meshes on a single GPU, achieves state-of-the-art accuracy on PDE surrogate benchmarks, and outperforms general-purpose efficient-attention methods on the Long Range Arena suite.
We additionally release a large-scale additive manufacturing benchmark dataset.
Our code is available at \url{https://github.com/vpuri3/FLARE.py}
\end{abstract}

\section{Introduction}

High-fidelity simulations of physical systems are often too costly for multi-query applications such as design optimization and uncertainty quantification.
Neural surrogate models offer a promising alternative by learning mappings from simulation inputs to outputs, enabling rapid evaluation once trained.

Transformers~\citep{vaswani2017attention} have demonstrated strong scalability and generalization across domains including natural language processing~\citep{devlin-etal-2019-bert} and computer vision~\citep{dosovitskiy2020image}.
This success has motivated their adoption for spatially distributed scientific data such as point clouds and unstructured meshes, where each mesh point is treated as a token with geometric and physical features.
However, standard self-attention scales as $\mathcal{O}(N^2)$ in both time and memory for $N$ tokens, which becomes prohibitive for high-resolution meshes containing hundreds of thousands to millions of points.


A common strategy for reducing attention cost is to introduce a low-dimensional latent bottleneck of size $M \ll N$, reducing complexity to $\mathcal{O}(NM)$.
Low-rank attention methods such as Linformer~\citep{wang2020linformer} assume that attention matrices are approximately low-rank and learn explicit $M\times N$ projection matrices\textbf{.}
While effective at moderate sequence lengths, this approach requires $\mathcal{O}(NM)$ parameters per layer and implicitly assumes a fixed token ordering, making it impractical for very long or unordered sequences such as unstructured meshes.
Latent-token architectures such as Perceiver and PerceiverIO~\citep{jaegle2021perceiver,jaegle2021perceiverio}, Transolver \citet{wu2024transolver}, and Latent Neural Operator (LNO)~\citet{wang2024latent} instead use attention projectors to map inputs to a fixed latent array, followed by deep latent-space self-attention.
Notably, most latent-token methods introduce additional transformations inside the latent space, making the overall attention behavior a composition of projections, latent mixing, and nonlinear refinements rather than a single, interpretable operator.

\begin{figure*}[t]
    \centering
    \includegraphics[width=0.9\linewidth, trim={0pt 10pt 0pt 0pt}]{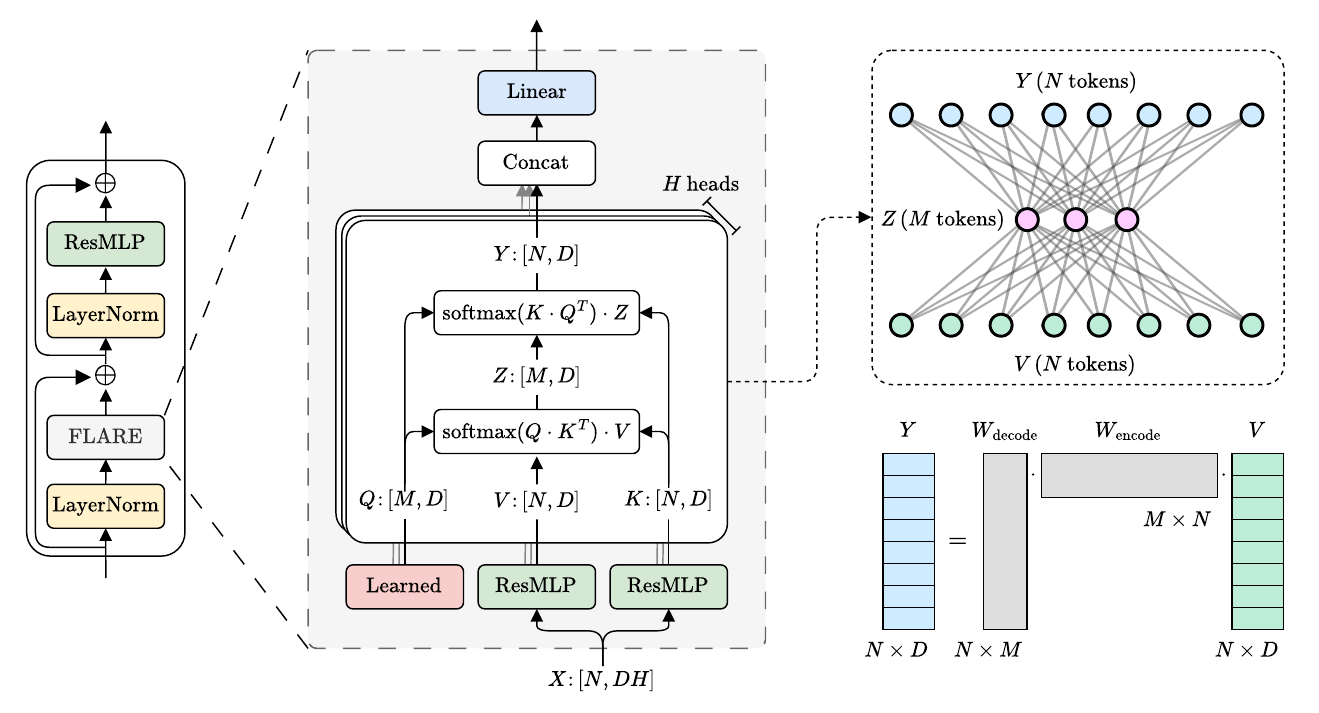}
    \caption{
    \textbf{FLARE block.}
    Each head encodes $N$ input tokens into $M$ latent tokens with cross-attention $W_{\mathrm{enc}}=\softmax(Q\cdot K^T)$ and decodes back with $W_{\mathrm{dec}}=\softmax(K\cdot Q^T)$, inducing an explicit rank-$\le M$ input-space mixing operator $W = W_{\mathrm{dec}}\cdot W_{\mathrm{enc}}$.
    Unlike PerceiverIO~\citep{jaegle2021perceiverio}, Transolver~\citep{wu2024transolver}, and LNO~\citep{wang2024latent}, FLARE uses no latent self-attention and executes both steps with fused SDPA kernels~\citep{dao2022flashattention} (see \autoref{fig:att_comparison} for a summary).
    }
    \vspace{-1\baselineskip}
    \label{fig:FLARE}
\end{figure*}

\begin{figure*}[t]
    \centering
    \includegraphics[width=1.0\linewidth]{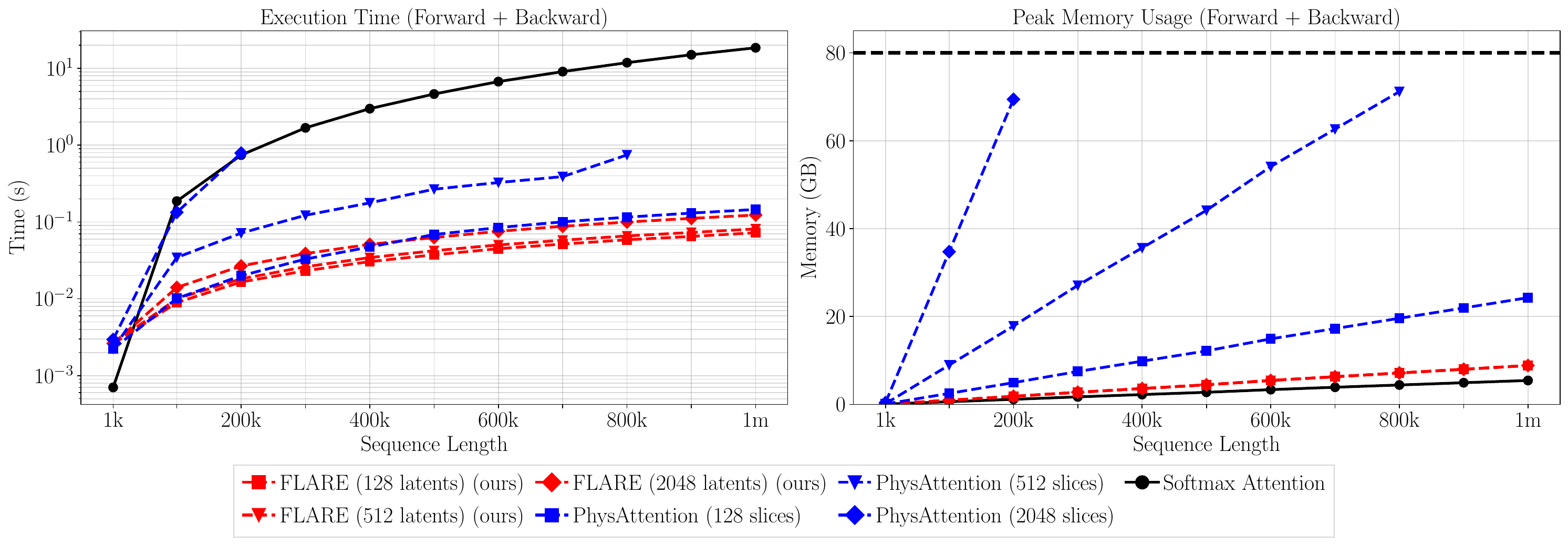}
    \caption{
    Time and memory requirements of different attention schemes.
    On an input sequence of one million tokens, FLARE (red) is over $200\times$ faster than vanilla attention, while consuming marginally more memory.
    All models are implemented with FlashAttention \citep{dao2022flashattention}, and
    the memory upper bound on a single H100 80GB GPU is depicted with a dashed line.
    Note that the curves for FLARE are somewhat overlapping.
    A detailed analysis is presented in \autoref{sec:appendix_ablation}.
    }
    \label{fig:time_memory}
\end{figure*}

We introduce \textbf{Fast Low-rank Attention Routing Engine (FLARE)}, a token-mixing layer that realizes a \emph{flexible low-rank self-attention operator} by routing information through a small set of latent tokens.
Concretely, each layer induces an input-input token mixing matrix of rank at most $M$ via an \emph{encode-decode} factorization implemented \emph{only} with two standard scaled dot-product attention (SDPA) calls; no additional token-mixing is performed.
This formulation makes the induced operator directly analyzable (e.g., via spectra) and keeps the dominant $\mathcal{O}(NM)$ computation fully SDPA-compatible, enabling fused attention kernels without materializing $M\times N$ projection weights.

FLARE operationalizes this idea with a minimal architecture: each block performs exactly two cross-attention operations (encode and decode).
Relative to Linformer~\citep{wang2020linformer}, FLARE parameterizes low-rank token mixing without learned $M\times N$ projection matrices or a fixed token ordering.
Relative to latent-token architectures such as PerceiverIO~\citep{jaegle2021perceiverio}, Transolver~\citep{wu2024transolver}, and LNO~\citep{wang2024latent}, which use latents as a computational workspace and introduce additional latent token mixing, FLARE uses latents only for routing: all token mixing is realized by a single encode-decode factorization on the input tokens, eliminating latent-space self-attention and yielding a mathematically explicit attention operator.
Counterintuitively, we find through systematic ablation that this architectural simplification (removing latent self-attention entirely) consistently improves accuracy while reducing parameters  compared to prior latent-attention methods.

FLARE further assigns disjoint latent slices to each attention head, yielding a mixture of head-specific low-rank operators rather than a single shared bottleneck.
As a result, each head implements an independent low-rank projection–reconstruction pathway, and global communication arises from aggregating multiple such operators.
We analyze these operators using a linear-time eigenanalysis method and observe diverse eigenvalue decay patterns across heads.
A controlled shared-latent ablation collapses these spectra and consistently degrades accuracy, indicating that independent latent routing is an important contributor to expressivity.
Together, these results support the interpretation that FLARE benefits from combining complementary low-rank pathways rather than relying on a single shared projection.

Taken together, the architectural features of FLARE, (i) explicit encode-decode low-rank factorization, (ii) elimination of latent-space self-attention, and (iii) head-wise independent latent routing, emerge from systematic empirical study as key principles for efficient, scalable attention.
This minimal design enables FLARE to achieve state-of-the-art performance on PDE surrogate benchmarks, scale to one-million-point unstructured meshes on a single GPU, and outperform general-purpose efficient-attention methods, all while using fewer parameters than existing approaches.


\autoref{fig:FLARE} illustrates a single FLARE block.
We summarize our main contributions below.

\begin{itemize}
\item \textbf{Flexible low-rank attention via latent routing.}
We introduce a self-attention formulation in which routing through latent tokens induces an implicit rank-$\le M$ attention operator on the input sequence.
FLARE eliminates all latent-space self-attention and expresses attention as a single encode-decode factorization, yielding a mathematically transparent operator.

\item \textbf{Head-wise independent low-rank pathways.}
By assigning each attention head its own latent slice, FLARE learns a mixture of parallel low-rank operators; targeted ablations show this head-wise independence is critical for accuracy.

\item \textbf{Extreme scalability with strong empirical performance.}
FLARE trains end-to-end on one-million-point unstructured meshes on a single GPU, achieves state-of-the-art PDE surrogate accuracy, and outperforms efficient-attention baselines on Long Range Arena.

\item \textbf{Benchmark dataset for additive manufacturing.}
We release a large-scale thermomechanical additive manufacturing dataset to support research on scalable scientific surrogates.
The dataset is available at \url{https://huggingface.co/datasets/vedantpuri/LPBF_FLARE}.
\end{itemize}

\section{Related work}

\paragraph{Neural PDE surrogates.}
Neural operators~\citep{li2020fourier,lu2021learning,kovachki2023neural} learn mappings between function spaces and enable mesh-independent generalization.
Extensions incorporating geometric priors and graph-based representations improve performance on unstructured meshes~\citep{li2023fourier,li2023geometry}.
Transformer-based architectures have emerged more recently as powerful PDE surrogates \citep{alkin2024universal, cao2021choose, li2022transformer, li2023scalable, hao2023gnot}, allowing global context aggregation.
Recent works \citep{alkin2024universal, wu2024transolver, luotransolver++, wang2024latent} leverage latent space attentions for PDE modeling, achieving high accuracy with reduced computational cost.
Building upon these advances, FLARE presents a linear-complexity attention operator tailored to extreme-resolution unstructured domains.

\paragraph{Efficient attention mechanisms.}
Efficient attention methods reduce cost by constraining attention structure (e.g., rank, sparsity, or locality), yielding restricted subsets of full self-attention with bounded expressivity.
Linformer~\citep{wang2020linformer} reduces attention complexity by learning explicit low-rank projections, but requires $\mathcal{O}(NM)$ parameters per layer.
PerceiverIO~\citep{jaegle2021perceiverio} and related models introduce latent tokens to reduce computational cost, using latent-space self-attention as a computational workspace to aggregate and transform global information.
In this design, interactions between input tokens are mediated through a sequence of latent-space transformations rather than defined by a single, explicit operator on the input tokens.
Transolver~\citep{wu2024transolver, luotransolver++} and LNO~\citep{wang2024latent} adapt this latent projection-unprojection paradigm for PDE surrogate modeling.
FLARE differs from these approaches by using latent tokens solely to induce an explicit low-rank operator, eliminating latent self-attention entirely and enabling a simpler, more analyzable attention formulation.



\section{Method}
\label{sec:method}

\subsection{Preliminary: Multi-Head Self-Attention}

Let $X \in \mathbb{R}^{N \times C}$ denote the input sequence of $N$ tokens with $C$ features each.
The query, key, and value matrices $Q, K, V \in \mathbb{R}^{N \times C}$ are obtained by applying learned linear projections to $X$:
\begin{equation}
    Q = X \cdot W^q,\hspace{1em}
    K = X \cdot W^k,\hspace{1em}
    V = X \cdot W^v,
\end{equation}
where $W^q, W^k, W^v\in\R^{C \times C}$.
The $Q, \, K, \, V$ matrices are then split along the feature dimension and passed to $H$ heads, each with dimension $D = C / H$, enabling parallel computation of attention:
\eqn{
\left[Q_1, \ldots, Q_H \right] = Q,\,
\left[K_1, \ldots, K_H \right] = K,\,
\left[V_1, \ldots, V_H \right] = V.
}
The scaled dot-product attention (SDPA) operation, introduced by \citep{vaswani2017attention}, computes the output as
\begin{equation}
\label{eq:sdpa}
Y_h = \text{SDPA}(Q_h, K_h, V_h, s) =
\softmax \left(\frac{Q_h \cdot K_h^T}{s}\right) \cdot V_h,
\end{equation}
where $Q_h,\, K_h, \, V_h \in \R^{N \times D}$ are query, key, and value matrices belonging to head $h$, and
$s$ is typically $\sqrt{D}$.
Note that $\softmax$ is taken along the row-dimension.
The outputs $Y_h$ from all heads are then concatenated along the feature dimension to form the final output
\begin{equation}
    Y = \left[ Y_1, \ldots, Y_H \right].
\end{equation}
This concatenation, followed by a linear layer, enables the model to integrate information across attention heads efficiently.

The greatest cost in multi-head self-attention is the call to SDPA which is $\O(N^2)$ in time and memory complexity.
This is because $Q_h \cdot K_h^T \in \R^{N \times N}$ requires $\O(N^2)$ storage and softmax, matrix-vector product with $V_h$ takes $\O(N^2)$ operations.
Fortunately, GPU-optimized multi-head implementations of SDPA are available in PyTorch \citep{paszke2019pytorch}.

\subsection{FLARE: Fast Low-rank Attention Routing Engine}

FLARE is a linear-complexity token mixing layer that learns low-rank global communication structures via attention projections.
The FLARE mechanism introduces a set of $M \ll N$ learnable latent tokens that serve as a bottleneck for information exchange.
The process consists of two stages:

\begin{enumerate}
\item
\textbf{Encoding.}
The input sequence is projected onto the latent tokens via cross-attention, compressing global information.
\item
\textbf{Decoding.}
The latent tokens are then projected back onto the input sequence, distributing the aggregated information.
\end{enumerate}

Formally, we define a learnable query matrix $Q \in \mathbb{R}^{M \times C}$, where each row corresponds to a latent token.
The key and value matrices, $K, V \in \mathbb{R}^{N \times C}$, are obtained by applying deep residual multi-layer perceptrons (MLPs) detailed in \autoref{sec:appendix_arch} to the input $X$.
Compared to just a linear layer, these allow the model to learn higher-order feature interactions and deeper nonlinear transformations. Refer to \autoref{sec:appendix_ablation} for ablation studies.

The matrices $Q$, $K$, and $V$ are first split along the feature dimension into $H$ heads, each of dimension $D = C / H$.
Then, for encoding, each head performs SDPA with a scaling factor $s=1$:
\begin{equation}
\label{eq:flare_encode}
Z_h = \mathrm{SDPA}(Q_h, K_h, V_h, s=1).
\end{equation}
Here, $Q_h \in \mathbb{R}^{M \times D}$, $K_h, V_h \in \mathbb{R}^{N \times D}$ are query, key, and value matrices belonging to head $h$ and $Z_h \in \mathbb{R}^{M \times D}$ is the latent sequence for head $h$.
For decoding and propagating information back to the input tokens, we perform a second SDPA operation, swapping the roles of queries and keys and using the latent sequence as values:
\begin{equation}
\label{eq:flare_decode}
Y_h = \mathrm{SDPA}(K_h, Q_h, Z_h, s=1)
\end{equation}
where $Y_h \in \mathbb{R}^{N \times D}$ is the output for each head.
Similar to multi-head self-attention, 
the outputs from all heads are concatenated along the feature dimension and passed through a final linear projection to mix information across heads.
As the query matrix has only $M$ tokens, the cost of $\text{SDPA}$ calls in \autoref{eq:flare_encode} and \autoref{eq:flare_decode} is $\O(NM)$.
PyTorch code for the multi-head implementation is presented in \autoref{fig:FLARE_pseudocode1}.
Note that we use a scaling factor $s=1$ instead of $\sqrt{D}$ in typical transformers \citep{vaswani2017attention} in SDPA.
This modification is explained in \autoref{sec:appendix_ablation}.

\begin{figure}[h]
\centering
\begin{lstlisting}[language=Python]
from torch.nn.functional import scaled_dot_product_attention as SDPA
def flare_multihead_mixer(Q, K, V):
    # Args - Q: [H M D], K, V: [B H N D]
    # Ret - Y: [B H N D]
    Z = SDPA(Q, K, V, scale=1.0)
    Y = SDPA(K, Q, Z, scale=1.0)
    return Y
\end{lstlisting}
\caption{
PyTorch code for multi-head token mixing operation in FLARE.
See \autoref{fig:FLARE_pseudocode2} for an implementation without the fused attention kernel.
}
\label{fig:FLARE_pseudocode1}
\end{figure}

\paragraph{Low-rank communication.}
The two-step attention process can be written as
\begin{equation}
Y_h = \left( W_{\text{decode},h} \cdot W_{\text{encode},h} \right) \cdot V_h
\end{equation}
where
\eqn{
W_{\text{encode},h} &= \mathrm{softmax}(Q_h \cdot K_h^T) \in \mathbb{R}^{M \times N},\quad \text{and}\\
W_{\text{decode},h} &= \mathrm{softmax}(K_h \cdot Q_h^T) \in \mathbb{R}^{N \times M}.
}
Note that $\softmax$ is taken along the row-dimension.
We define
\eqn{
    \label{eq:attn_W}
    W_h = W_{\text{decode},h} \cdot W_{\text{encode},h} \in \mathbb{R}^{N \times N}
}
as the dense global communication matrix with rank at most $M$.
This low-rank structure, illustrated in \autoref{fig:FLARE}, enables efficient all-to-all communication without explicitly forming $W_h$; instead, $W_{\text{encode},h}$ and $W_{\text{decode},h}$ are applied sequentially, at an overall cost of $\mathcal{O}(MN)$ per head.

\begin{table*}[t]
\centering
\caption{
(Top) relative $L_2$ error ($\times 10^{-3}$) and (bottom) parameter count for different models across PDE benchmark problems.
The best results (smallest error) are made bold, and the second best results are underlined.
A backslash ($\backslash$) indicates that the model cannot be applied to the benchmark,
and tilde ($\sim$) indicates that the model is prohibitively slow on the benchmark.
}
\label{tab:pde_benchmarks}
\footnotesize
\begin{tabular}{l|cccccc}
\toprule
\textbf{Model} &
\makecell[c]{\textbf{Elasticity}   } &
\makecell[c]{\textbf{Darcy}        } &
\makecell[c]{\textbf{Airfoil}      } &
\makecell[c]{\textbf{Pipe}         } &
\makecell[c]{\textbf{DrivAerML-40k}} &
\makecell[c]{\textbf{LPBF}         } \\
\midrule
\makecell[l]{Vanilla Transformer \\ \citep{vaswani2017attention}} &
\makecell{    \ul{5.37} \\ 660k} & 
\makecell{\textbf{4.38} \\ 660k} & 
\makecell{       {6.28} \\ 660k} & 
$\sim$                           & 
$\sim$                           & 
$\sim$                           \\ 
\midrule
\makecell[l]{{PerceiverIO} \\ \citep{jaegle2021perceiverio}} &
\makecell{{28.0} \\ {1.87m}} & 
\makecell{{20.6} \\ {1.87m}} & 
\makecell{{7.65} \\ {1.87m}} & 
\makecell{{6.90} \\ {1.87m}} & 
\makecell{{248 } \\ {1.87m}} & 
\makecell{{23.1} \\ {1.87m}} \\ 
\midrule
\makecell[l]{GNOT \\ \citep{hao2023gnot}} &
\makecell{13.3 \\ 4.87m} & 
\makecell{16.9 \\ 4.90m} & 
\makecell{103  \\ 4.90m} & 
\makecell{5.89 \\ 4.90m} & 
\makecell{115  \\ 4.87m} & 
\makecell{24.3 \\ 4.87m} \\ 
\midrule
\makecell[l]{LNO \\ \citep{wang2024latent}} &
\makecell{9.25 \\ 1.83m} & 
\makecell{7.64 \\  762k} & 
\makecell{17.8 \\  762k} & 
\makecell{8.10 \\  762k} & 
\makecell{146  \\  762k} & 
\makecell{24.7 \\  762k} \\ 
\midrule
\makecell[l]{Transolver w/o conv\\ \citep{wu2024transolver}} &
\makecell{    6.40  \\ 713k} & 
\makecell{    18.6  \\ 713k} & 
\makecell{    8.24  \\ 713k} & 
\makecell{    4.87  \\ 713k} & 
\makecell{\ul{70.5} \\ 713k} & 
\makecell{\ul{20.4} \\ 713k} \\ 
\midrule
\makecell[l]{Transolver with conv\\ \citep{wu2024transolver}} &
$\backslash$                 & 
\makecell{   {5.94} \\ 2.8m} & 
\makecell{\ul{5.50} \\ 2.8m} & 
\makecell{\ul{3.90} \\ 2.8m} & 
$\backslash$                 & 
$\backslash$                 \\ 
\midrule
\textbf{FLARE (ours)} &
\makecell{\textbf{3.38} \\ 592k} & 
\makecell{    \ul{5.10} \\ 691k} & 
\makecell{\textbf{4.28} \\ 691k} & 
\makecell{\textbf{2.85} \\ 625k} & 
\makecell{\textbf{60.8} \\ 691k} & 
\makecell{\textbf{18.5} \\ 625k} \\ 

\bottomrule
\end{tabular}
\end{table*}

\paragraph{Discussion on the design principles of FLARE.}
FLARE's latent tokens implement a structured gather-scatter communication pattern: encoding pools input values into $M$ compact global descriptors, then decoding selectively broadcasts these back along the same pathways, yielding a low-rank routing operator that avoids explicit all-to-all interactions.
This mechanism relies on two design choices discussed in \autoref{sec:appendix_flare_design}: (i) symmetric encode-decode operators promote stable information flow without redundant parameters, and (ii) fixed input-independent latent queries simplify the low-rank structure and improve efficiency, shifting expressive capacity to deep residual key/value projections that recover representational power.


\paragraph{FLARE block.}
\autoref{fig:FLARE} (left) illustrates a single FLARE block.
Given input tokens $X\in\R^{N \times C}$, the output of an FLARE block is computed as
\eqn{
    X &= X + \text{FLARE}\left( \text{LayerNorm}\left( X \right) \right)\\
    X &= X + \text{ResMLP}\left( \text{LayerNorm}\left( X \right) \right).
}
Here, $\text{ResMLP}$ \citep{he2016deep} denotes a residual MLP block detailed in \autoref{sec:appendix_arch}, and $\text{LayerNorm}$ denotes layer normalization \citep{ba2016layer} operation.
To summarize, a FLARE block consists of a token mixing operation via FLARE, a pointwise residual MLP, and layer normalization in pre-norm format~\citep{xiong2020layer}.
Deep residual MLPs within the block enable complex, token-level feature transformations and improve accuracy.

\paragraph{Overall design.}
The overall architecture is given by $B$ sequential FLARE blocks sandwiched between an input projection and an output projection, which are detailed in \autoref{sec:appendix_arch}.
Such a design enables the model to efficiently integrate local and global information across multiple layers, making it well suited for large-scale, high-dimensional data such as point clouds.

\begin{figure*}[t]
\centering

\begin{subfigure}[t]{\linewidth}
    \centering
    \includegraphics[width=0.48\linewidth, trim={0 3pt 0pt 0pt}]{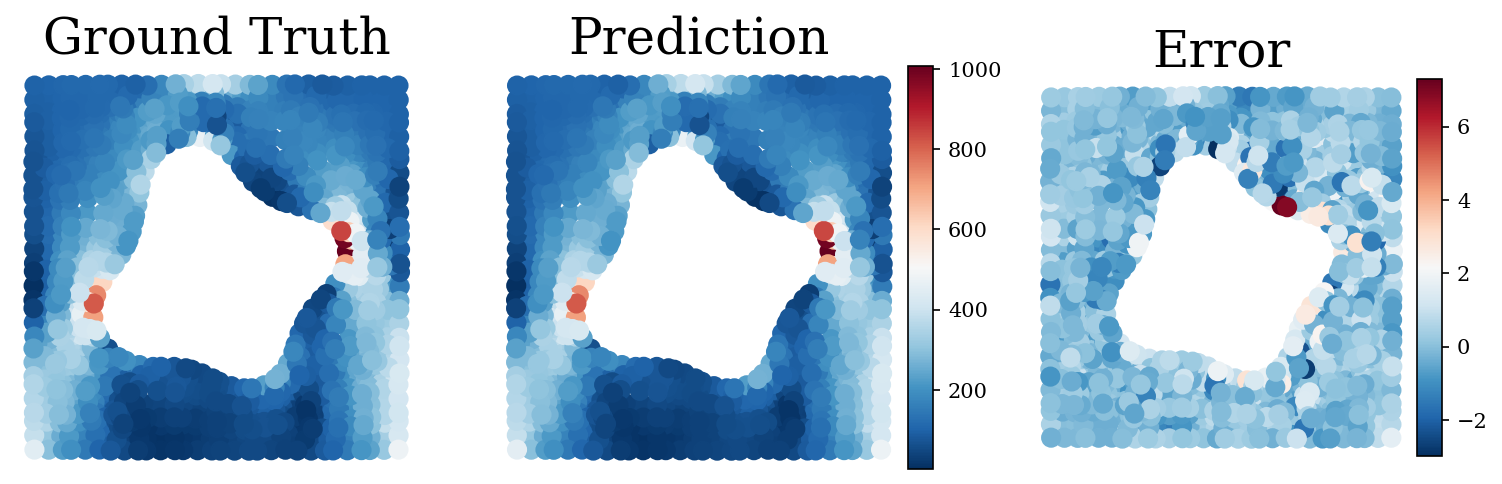}
    \includegraphics[width=0.48\linewidth, trim={0 0pt 0pt 3pt}]{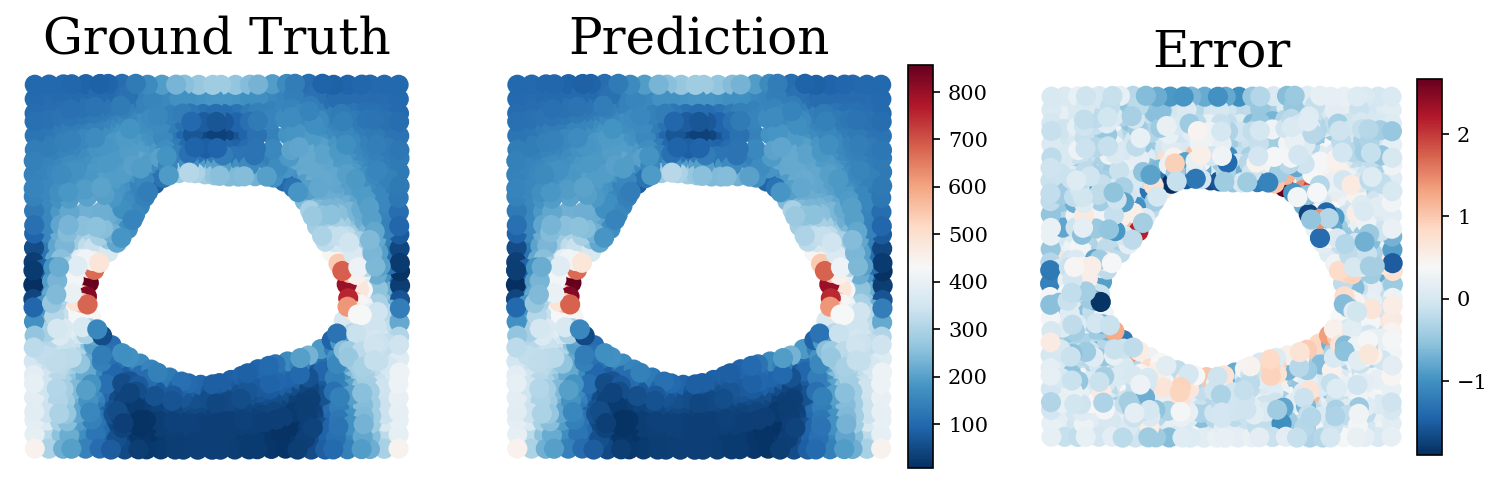}
    \hfill
    \caption{Elasticity}
\end{subfigure}

\begin{subfigure}[t]{0.48\linewidth}
    \centering
    \includegraphics[width=\linewidth, trim={0 0pt 0pt 3pt}]{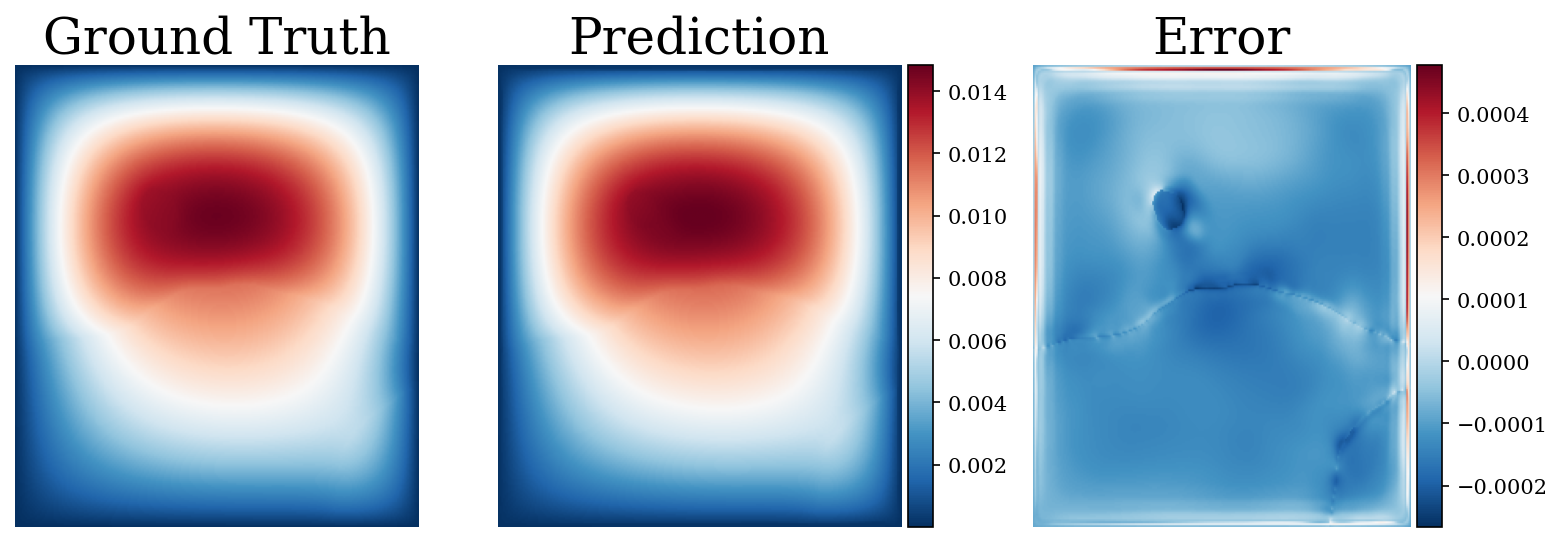}
    \caption{Darcy}
\end{subfigure}
\hfill
\begin{subfigure}[t]{0.48\linewidth}
    \centering
    \includegraphics[width=\linewidth, trim={0 0pt 0pt 3pt}]{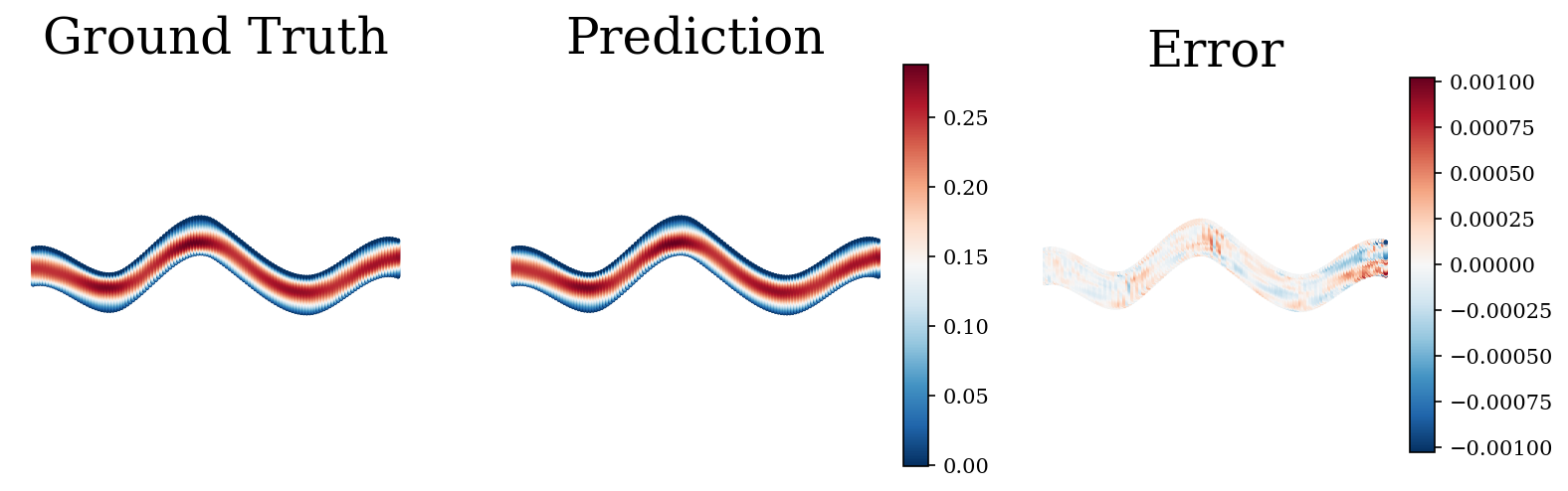}
    \caption{Pipe}
\end{subfigure}

\caption{
Qualitative results for FLARE on the (a) Elasticity, (b) Darcy, and (c) Pipe datasets.
We show the ground truth, the model prediction, and the corresponding error (Ground Truth $-$ Prediction).
}
\label{fig:flare_results}
\end{figure*}

\subsection{Spectral analysis}
\label{sec:spectra}

The matrix $ W_h $ in \autoref{eq:attn_W} represents the attention weights between $ N $ tokens, where $ [W_h]_{ij} $ quantifies how much token $j$ communicates to token $i$ within head $h$.
Since $W_h$ has rank at most $M$, the model can capture at most $M$ independent global communication patterns per head.
The eigenvalues of $ W $ indicate the relative importance or energy of each latent dimension in forming the attention matrix $ W $.
We apply an algorithm to obtain the eigendecomposition of $W$ in $\mathcal{O}(M^3 + M^2N)$ time compared to $\mathcal{O}(N^3)$ for a dense communication matrix.
The algorithm is predicated on computing the eigenspectra of an $M \times M$ matrix $JJ^T$
where $J \in \R^{M \times N}$ is chosen such that $J^T J$ is similar to $W$.
The algorithm is detailed in \autoref{sec:appendix_spectral_method}, and summarized in Algorithm 1. 
\autoref{fig:spectra} presents the $M$ nonzero eigenvalues of $W_h$ for an FLARE model trained on the Elasticity benchmark problem with 972 points per input.
The distinct spectra of the heads indicates that each head learns distinct attention patterns.

The eigenvalue analysis detailed in \autoref{sec:appendix_spectra_qualitative} shows that while FLARE provides capacity for rank-$M$ attention, the model learns to use only a small fraction of this in early blocks indicating effective compression.
In deeper blocks, more of the latent capacity is utilized, with diverse spectral profiles across heads, validating our design choice of independent head-wise projections.

\section{Benchmark dataset for additive manufacturing}

We introduce a benchmark dataset of \emph{laser powder bed fusion} (LPBF) simulations designed to evaluate surrogate models on large, irregular 3D geometries.
LPBF is a widely used metal additive manufacturing process in which a laser fuses thin layers of powder, producing complex parts but often inducing \emph{residual stresses} and \emph{distortions} that can lead to build failures \citep{zhang_flaw_nodate}.
To capture these effects, we simulate the thermo–mechanical LPBF process on thousands of geometries drawn from the Fusion~360 segmentation dataset \citep{lambourne2021brepnet}.

Our benchmark task is \emph{predicting vertical ($Z$) displacement field} for each geometry, a quantity directly associated with recoater-blade collisions and build-failure risk in LPBF.
Each sample consists of the 3D mesh coordinates as input and the final $Z$-displacement at all nodes as output.
The dataset spans a wide variety of part shapes, mesh resolutions, and deformation magnitudes, making it a challenging and practically meaningful testbed for large-scale surrogate modeling.
Additional details such as statistics and visualizations are provided in \autoref{sec:appendix_am_dataset}.

\section{Experiments}

\subsection{PDE surrogate benchmarks}
\label{sec:experiments_pde_surrogates}

\paragraph{Benchmark problems.}


We consider a diverse set of benchmark datasets (\autoref{tab:dataset_summary}) for regressing PDE solutions on point clouds spanning structured and unstructured grids with up to 50,000 points.
Note that FLARE is mesh-agnostic, and operates solely on the input point cloud.
The 2D Elasticity, Darcy, Airfoil, and Pipe benchmarks \citep{li2020fourier, li2023fourier} cover a wide range of physical phenomena, and the 3D DrivAerML benchmark \citep{ashton2024drivaerml} provides automotive aerodynamic simulations.
Visualizations of the Elasticity, Darcy, and Pipe benchmark problems are presented in \autoref{fig:flare_results}.
Additional details of the dataset are presented in \autoref{sec:benchmark_datasets}.
We also introduce a 3D field-prediction benchmark derived from laser powder bed fusion (LPBF) simulations, with diverse 3D-printed parts containing up to 50,000 grid points (see \autoref{sec:appendix_am_dataset}).

\begin{figure*}[t]
    \centering
    \includegraphics[width=1.0\linewidth]{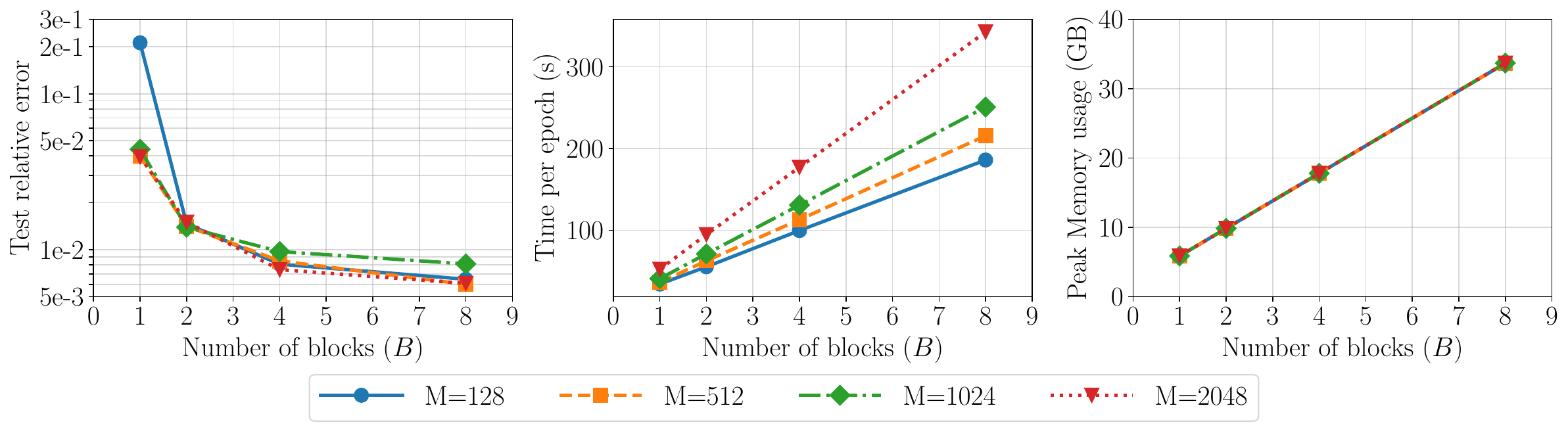}
    \caption{
    We train FLARE on the DrivAerML dataset \citep{ashton2024drivaerml} with one million points per geometry on a single Nvidia H100 80GB GPU.
    We present (left) the test relative error, (middle) time per epoch (s), and
    (right) peak memory utilization (GB) as a function of the number of FLARE blocks ($B$) for different number of latent tokens ($M$).
    }
    \label{fig:scale_dml}
\end{figure*}

\paragraph{Baselines.}
We compare FLARE with state-of-the-art PDE surrogates: generic attention models (vanilla Transformer \citep{vaswani2017attention}, PerceiverIO \citep{jaegle2021perceiverio}); attention-based PDE surrogates (Transolver \citep{wu2024transolver}, LNO \citep{wang2024latent}); and the neural operator GNOT \citep{hao2023gnot}.
We exclude graph-based models because graph connectivity is unavailable for most problems.
Transolver++ \citep{luotransolver++} is a concurrent work; we tested the official implementation with reported hyperparameters, but were unable to reproduce the reported numbers, so we do not include it in our comparisons (details in \autoref{sec:appendix_benchmark_models}).
We follow the experimental setup of Transolver as it is the preeminent surrogate model and attempt to match its parameter count.
Note that Transolver can be instantiated in two configurations: \emph{without convolution}, where point-to-point communication relies solely on physics attention,
and \emph{with convolution}, where convolution layers are added to inject information from neighboring points when the input grid is structured.
We evaluate these two configurations separately to isolate the impact of convolution versus physics attention.
In our model, we choose not to employ any convolution layers and rely entirely on FLARE for token mixing. Because these PDE problems are relatively small (up to 50,000 points), we train all models in FP32. As shown in \autoref{fig:time_memory_fp32}, the vanilla Transformer is drastically slower than FLARE and Transolver on large point clouds; accordingly, we evaluate it only on problems up to \( \sim 10{,}000 \) points.

\paragraph{Discussion.}
The results in \autoref{tab:pde_benchmarks} clearly demonstrate that the proposed FLARE architecture achieves the lowest relative $L_2$ error across all but one benchmark PDE problems, outperforming both LNO and Transolver on every dataset.
Notably, FLARE also achieves these gains with a consistently lower parameter count than Transolver or LNO, highlighting its efficiency in addition to higher accuracy.
These results underline the robustness and versatility of FLARE across diverse problem settings.
We also note that the poor performance of Transolver without convolutions indicates that the inter-point communication via Transolver's built-in physics-attention mechanism is not enough.
With convolutions, the input projections amass information from neighboring points, which in turn helps the physics attention learn the global structure.

Although the vanilla transformer is extremely effective on small-scale PDE problems, it becomes prohibitively slow on large point clouds due to its quadratic cost as illustrated in \autoref{fig:time_memory}.
On the contrary, PerceiverIO (with only a single encoding and decoding step) performs poorly even with $M=1,024$ latent tokens and $B=8$ latent self-attention blocks.
This validates our hypothesis that
multiple latent self-attention operations can be unnecessary and potentially suboptimal so long as the projections are sufficiently expressive.
This is because information loss during projection is not recoverable via latent self-attention alone.
Instead, performing multiple (head-wise) parallel projections and reconstructions directly between the input and latent sequences preserves expressivity while simplifying the architecture.

\begin{table*}[t]
\centering
\caption{
Accuracy (\%) of different transformer models on Long Range Arena benchmark tasks \cite{taylong}.
The best result (highest accuracy) is \textbf{bold} and the second best is \underline{underlined}.
}
\label{tab:experiments_lra}
\begin{tabular}{l|cccccc}
\toprule
Model & \textbf{ListOps} & \textbf{Text} & \textbf{Retrieval} & \textbf{Image} & \textbf{Pathfinder-32} & \textbf{Avg} \\
\midrule
Vanilla attention 
    & \textbf{36.70}
    & \underline{64.93}
    & \underline{77.18}
    & 38.02
    & 70.52
    & \underline{57.47} \\
Linear attention 
    & 17.15
    & \textbf{66.00}
    & 71.84
    & 09.86
    & \textbf{75.00}
    & 47.97 \\
Linformer 
    & \textbf{36.70}
    & 53.00
    & 64.72
    & \textbf{41.88}
    & 70.09
    & 53.28 \\
Norm attention 
    & 17.10
    & 63.08
    & 76.07
    & 36.94
    & 70.15
    & 52.67 \\
Performer 
    & 35.90
    & 64.21
    & 68.42
    & 35.36
    & 53.83
    & 51.54 \\
\textbf{FLARE (ours)} 
    & \underline{36.15}
    & 64.00
    & \textbf{77.30}
    & \underline{40.96}
    & \underline{71.91}
    & \textbf{58.06} \\
\bottomrule
\end{tabular}
\end{table*}

\subsection{Field-prediction on million-point geometries}

The benchmark problems in \autoref{sec:experiments_pde_surrogates} contain a variety of PDE problems, but are relatively small compared to industrial use cases that demand PDE solutions on complex geometries with millions of grid points \citep{ashton2024drivaerml}.
So far, attention-based surrogate models have not been able to scale to million-scale regression problems due to quadratic time and memory complexity, as illustrated in \autoref{fig:time_memory}.
The FlashAttention \citep{dao2022flashattention} algorithm has alleviated the memory bottleneck thanks to online softmax computation; however, these methods remain impractical due to their long training times.
Furthermore, SOTA models such as Transolver and LNO are not readily expressible purely in terms of SDPA calls for their dominant $\mathcal{O}(NM)$ projection/unprojection steps; in typical implementations these steps explicitly materialize $M\times N$ projection weights, which prevents using PyTorch's fused SDPA backends (e.g., FlashAttention-style kernels) for the $\mathcal{O}(NM)$ bottleneck.

We demonstrate in \autoref{fig:scale_dml} that FLARE can scale to million-scale geometries by training on the DrivAerML dataset \citep{ashton2024drivaerml} where each mesh is subsampled to contain one million points.
These calculations are performed in mixed precision on a single Nvidia H100 80GB GPU provisioned through Google Cloud Platform.
We note the clear trend in \autoref{fig:scale_dml} (left) that the error consistently decreases as we increase the number of FLARE blocks.
To our knowledge, this is the first attention-based neural surrogate model trained on one million points on a single GPU without memory offloading or distributed computing.
Additional details are presented in \autoref{sec:appendix_field_pred}.

\subsection{Long Range Arena benchmark problems}
\label{sec:experiments_lra}

To demonstrate that FLARE is not limited to PDE surrogate modeling, we also evaluate it on the Long Range Arena (LRA) benchmark suite \citep{taylong}, which covers a diverse set of long-context tasks ranging from logical reasoning and text classification to retrieval, image classification, and visual pathfinding.
While many top-performing LRA models rely on architectures specialized for 1D sequence structure (e.g., linear-attention kernels and state-space models such as MEGA \citep{ma2022mega} and Mamba \citep{gu2024mamba}), these methods typically assume an inherent token ordering and are therefore not directly applicable to the unordered point clouds and unstructured meshes that motivate this work.
In contrast, FLARE is fully permutation-equivariant and makes no such structural assumptions.
As shown in Table~\ref{tab:experiments_lra}, FLARE achieves the highest average accuracy among the attention baselines we evaluate, outperforming general-purpose efficient-attention mechanisms (Linformer \citep{wang2020linformer}, Performer \citep{choromanski2020rethinking}, Norm Attention \citep{qin2022devil}, Linear Attention) and even vanilla self-attention on average.
These results indicate that the low-rank formulation underlying FLARE provides a robust and broadly applicable inductive bias, extending beyond PDE surrogates to heterogeneous long-range tasks.


\subsection{Model analysis and ablation studies}
\label{sec:experiments_model_analysis}

We conduct focused ablations in \autoref{sec:appendix_ablation} that causally isolate the contribution of each component and clarify trade-offs between expressivity, time, and memory complexity.
Runnable code for all experiments, benchmarks, and ablations is provided in the supplementary material.

We first examine \textbf{time and memory scaling} as a function of sequence length.
FLARE exhibits true linear $\mathcal{O}(NM)$ scaling in both runtime and peak memory usage, even up to one million tokens.
This behavior follows directly from FLARE’s SDPA-compatible formulation, which avoids materializing attention matrices or $M\times N$ projection weights and allocates sequence-length–dependent memory only for input activations.

Next, we study the trade-off between \textbf{model depth and latent dimensionality}.
Across all PDE benchmarks, including the 1M-point DrivAerML dataset, accuracy improves consistently with more encode–decode blocks, while increasing the number of latent tokens $M$ yields diminishing returns except on inherently high-rank tasks.
This demonstrates that repeated low-rank mixing through depth is a more effective driver of global communication than increasing the rank of a single attention operator.

We then evaluate \textbf{head-wise parallel low-rank pathways}.
Varying the number of attention heads while keeping total width fixed shows that FLARE performs best with many small heads rather than a few large ones, indicating that aggregating multiple independent low-rank operators approximates richer attention patterns than concentrating capacity into fewer projections.

Next, we compare \textbf{latent-space self-attention versus encode–decode blocks} and find that latent self-attention consistently worsens accuracy and increases computation, whereas adding encode–decode blocks improves performance; the best results occur when latent self-attention is entirely removed.
Then, we study \textbf{shared versus independent latent tokens} and observe that sharing latents across heads collapses the induced spectra and degrades accuracy, while independent latent slices yield diverse eigenvalue decay profiles and lower error.

Together, these ablations validate FLARE’s architectural choices: explicit low-rank operator factorization, repeated encode-decode mixing, and head-wise latent independence. Its gains, therefore, do not stem from increased parameter count or implementation-level optimizations.

\section{Conclusion}

FLARE is a token mixing layer that bypasses the quadratic cost of self-attention by leveraging low-rankness.
Mechanically, FLARE routes attention through a fixed-size latent sequence via cross-attention projection and unprojection.
FLARE achieves SOTA accuracy on a set of diverse PDE benchmarks, and easily scales to PDE problems with million-scale geometries.

As transformers are the backbone of modern deep learning, we postulate that an efficient attention mechanism has several applications.
We also identify potential areas for improvement:
FLARE’s reliance on deep residual MLPs can introduce sequential bottlenecks and increase latency, suggesting that further speedups are possible by addressing this issue.
Additional enhancements include
(1) incrementally increasing the number of latent tokens during training;
(2) conditioning the latent queries/tokens on the input to relax the fixed-$Q$ constraint;
(3) conditioning latent tokens on time for diffusion modeling;
and
(4) designing decoder-only variants for autoregressive modeling.
\section*{Acknowledgments}
This work was supported by the Air Force Research Laboratory under contract FA8650-21-F-5803 and PA Manufacturing Fellows Initiative. Zhang was supported in part by the National Science Foundation under grants CMMI-1953323 and CBET-2332084.
The authors appreciate the support of Camfer, Inc.~\citep{camfer2025} and Professor Amir Barati Farimani in providing computing resources.
Part of this research was conducted on the Bridges-2 Supercomputer at the Pittsburgh Supercomputing Center, and parts were conducted using ORCHARD, a high-performance cloud computing cluster made available by Carnegie Mellon University.
The authors thank Jay Pathak of Ansys, Inc.~\citep{ansys2025} for insightful discussions, undergraduate researcher Andrew Porco for his assistance with data processing.

\section*{Impact Statement}

Self-attention underpins many modern AI systems, but its quadratic computational and memory cost makes both training and inference expensive and difficult to scale.
By making attention-based models practical on very large inputs within single-GPU workflows, FLARE lowers the barrier to deploying powerful sequence models across a wide range of applications.

\bibliography{main}
\bibliographystyle{icml2026}

\newpage
\appendix

\onecolumn

\section{Extended related work}
\label{sec:appendix_related_work}

\begin{figure*}[t]
\centering
\includegraphics[width=1.0\linewidth, trim={0 -10pt 0pt 0pt}]{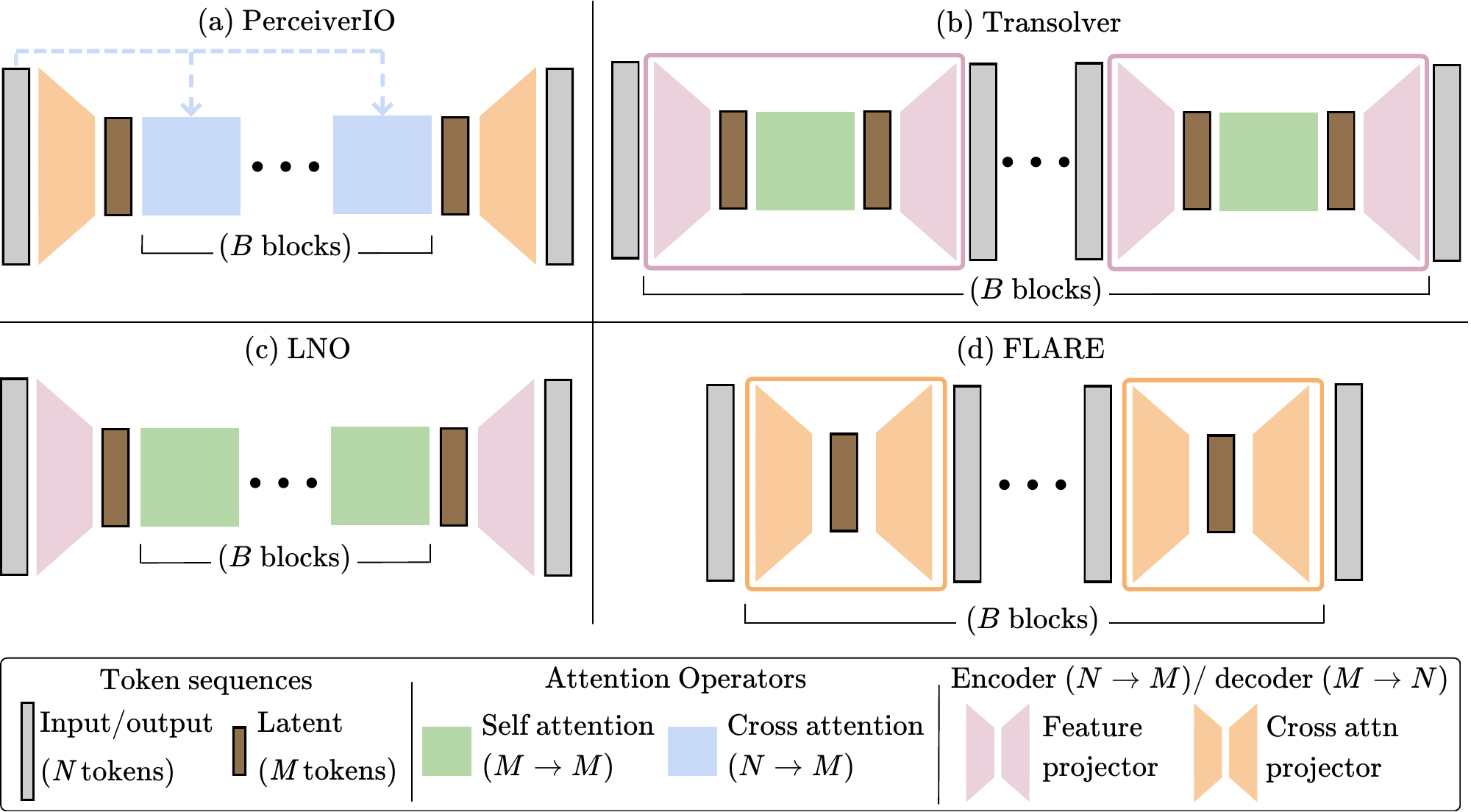}
\begin{threeparttable}
\footnotesize
\begin{tabular}{l|cccc}
\toprule
{Criteria} &
{PerceiverIO} &
{Transolver} &
{LNO       } &
{FLARE} \\
\midrule

\makecell[l]{Asymptotic evaluation cost*} &
\footnotesize $2MN + BMN$   &
\footnotesize $2BMN + BM^2$ &
\footnotesize $2MN + BM^2$  &
\footnotesize $2BMN$        \\

\makecell[l]{Sequential enc/dec blocks} &
\xmark &
\cmark &
\xmark &
\cmark \\

\makecell[l]{Headwise parallel enc/dec} &
\cmark &
\cmark$^\dagger$ &
\xmark &
\cmark \\

\makecell[l]{Encoder/decoder coupling} &
\xmark &
\cmark &
\cmark &
\cmark \\

\makecell[l]{Latent space attention} &
\cmark &
\cmark &
\cmark &
\xmark \\

\makecell[l]{Fused attention compat.} &
\cmark &
\xmark &
\xmark &
\cmark \\

\bottomrule
\end{tabular}
\begin{tablenotes}[flushleft]
\footnotesize
\item *Evaluating cost of processing $N \gg M$ tokens ignoring feedforward layers (FFN are $\O(N)$).
\item $^\dagger$Physics Attention in Transolver uses the same projection weights for all heads,
whereas PerceiverIO and FLARE use cross-attention projection where each head learns
a distinct slice of the latent tokens.
\end{tablenotes}
\caption{
Comparison of
(a) PerceiverIO \citep{jaegle2021perceiverio},
(b) Transolver \citep{wu2024transolver},
(c) LNO \citep{wang_latent_2024}, and
(d) FLARE (ours).
Each model contains $B$ blocks and $M$ latents.
The criteria in the table are described in detail in \autoref{sec:appendix_related_work}.
}
\vspace{-2\baselineskip}
\label{fig:att_comparison}
\end{threeparttable}
\end{figure*}

\autoref{fig:att_comparison} provides a structural and computational comparison between PerceiverIO \citet{jaegle2021perceiverio}, Transolver \citet{wu2024transolver}, Latent Neural Operator (LNO) \citet{wang2024latent}, and FLARE.
All four architectures introduce a latent bottleneck to reduce the quadratic cost of self-attention, but they differ fundamentally in how latent tokens are used, how information is mixed across tokens, and how the resulting attention operator is realized and implemented.

\paragraph{Perceiver and PerceiverIO.}
Perceiver and PerceiverIO~\citep{jaegle2021perceiver,jaegle2021perceiverio} introduce a general latent-attention framework in which a fixed-size latent array of length $M$ attends to an input sequence of length $N$ via cross-attention.
This step maps the input into latent space, after which multiple layers of latent self-attention are applied.
Optionally, PerceiverIO performs additional cross-attention from the latent array back to the input or output tokens.
This design reduces the cost of attention on the input sequence and enables flexible input and output modalities.

However, Perceiver-style models use latent tokens as a \emph{computational workspace}: information is repeatedly transformed within the latent space through self-attention.
As a result, the effective input-input communication pattern is a deep composition of latent transformations rather than a single, explicit attention operator.
While expressive, this structure makes it difficult to characterize the induced token-to-token interactions or to interpret the model as implementing a specific low-rank attention operator on the input sequence.

\paragraph{Transolver and Latent Neural Operator.}
Transolver~\citep{wu2024transolver} and Latent Neural Operator (LNO)~\citep{wang2024latent} adapt latent-attention ideas specifically for PDE surrogate modeling on point clouds and meshes.
Both architectures project variable-length inputs into a fixed-length latent representation and use latent self-attention to aggregate global information.
A key difference from PerceiverIO is that Transolver performs projection and unprojection in \emph{every} transformer block, allowing global context to accumulate progressively with depth.
LNO, by contrast, performs a single projection-unprojection step followed by latent self-attention.

In practice, both Transolver and LNO rely on feature-projection mechanisms that explicitly construct or manipulate $M\times N$ projection weights, typically via feature expansion followed by attention or linear layers.
These projection layers introduce additional memory and compute overhead that becomes significant for large $N$ and $M$.
Moreover, both models rely on latent-space self-attention as the primary mechanism for global mixing, which introduces an $\mathcal{O}(M^2)$ cost per block.

We note that Transolver can be instantiated with or without convolutional layers.
Without convolution, point-to-point communication relies solely on physics attention; with convolution, local neighborhood aggregation is explicitly injected before attention.
We evaluate both configurations separately to disentangle the effects of convolutional inductive bias and latent attention.
Our results in \autoref{tab:pde_benchmarks} show that Transolver without convolution performs poorly, indicating that its physics-attention mechanism alone is insufficient for effective global communication.
When convolutions are included, local aggregation significantly improves performance, suggesting that much of the gain arises from convolutional preprocessing rather than latent attention itself.
In contrast, FLARE does not rely on any convolutional layers and achieves strong performance using attention-based token mixing alone.

\paragraph{FLARE: latent routing as a low-rank operator.}
FLARE differs from prior latent-attention architectures in how latent tokens are used.
Rather than serving as a computational workspace, latent tokens in FLARE act purely as a \emph{routing mechanism}.
Each block consists of exactly two cross-attention operations (encode and decode) that together induce an explicit rank-$\le M$ linear operator on the input tokens.
FLARE eliminates latent-space self-attention entirely, ensuring that all token mixing occurs in the input space through a single encode-decode factorization.

A further distinction is FLARE’s use of \emph{head-wise independent latent slices}.
Each attention head projects into its own latent subspace, forming multiple parallel low-rank projection-reconstruction pathways.
This contrasts with Transolver, which shares projection weights across heads, and LNO, which uses a single global projection.
As shown in our spectral analysis and ablations, head-wise independence enables different heads to specialize in complementary low-rank routing patterns rather than collapsing into a shared subspace.

\paragraph{Summary of architectural trade-offs.}
As summarized in \autoref{fig:att_comparison}, FLARE combines several desirable properties of prior approaches while avoiding their key limitations.
Like Transolver, it performs encode-decode operations in every block, enabling progressive global mixing.
Like PerceiverIO, it relies on cross-attention and is fully compatible with fused SDPA kernels.
Unlike all three prior methods, FLARE eliminates latent self-attention and uses head-wise independent latent routing, yielding a simple, interpretable, and strictly linear-time attention operator.
These differences explain both FLARE’s scalability to extreme sequence lengths and its strong empirical performance.

Below we briefly clarify the criteria used in \autoref{fig:att_comparison}.
Asymptotic evaluation cost refers to the dominant attention-related complexity for processing $N \gg M$ tokens with $B$ blocks, ignoring feed-forward layers which are linear in $N$.
The remaining criteria describe architectural structure rather than cost.
Sequential encode-decode blocks indicate whether global context is progressively refined across depth (as in Transolver and FLARE).
Head-wise parallel encode-decode denotes whether each attention head learns an independent latent projection, which is true for PerceiverIO and FLARE but not for Transolver or LNO.
Encoder/decoder coupling refers to whether encoding and decoding projections are tied, and latent space attention indicates the presence of explicit self-attention over latent tokens.

Finally, fused-attention compatibility captures whether all attention operations can be expressed using standard scaled dot-product attention (SDPA) primitives.
PerceiverIO and FLARE satisfy this property, enabling efficient fused implementations, whereas Transolver and LNO rely on additional projection operations that limit hardware efficiency at large $N$.
Note that Transolver can employ fused SDPA kernels for latent-space self-attention ($\O(M^2)$). However, encoding and decoding ($\O(NM)$) operations, which form the bottleneck in these computations, are implemented with linear projections and cannot employ SDPA kernels.

\section{Architecture details}
\label{sec:appendix_arch}

\subsection{Input/ output projection}

\paragraph{ResMLP.}
We implement a residual MLP block to serve as a flexible non-linear function approximator.
Given input/output dimensions $C_\text{i}$ and $C_\text{o}$, the layer first applies a linear transformation to a hidden space of size $C_\text{h}$, followed by $L$ residual layers, each consisting of a linear layer with GELU activation \citep{hendrycks2016gaussian}.
These are the only instances of pointwise nonlinear activations in the model.
An optional input residual connection is applied after the first layer when $C_\text{i}=C_\text{h}$, and an optional output residual connection is applied at the end when
$C_\text{h}=C_\text{o}$.
The final output is projected to dimension $C_\text{o}$ via a linear layer.
This design allows control over depth and expressivity while preserving stability through residual connections.

\paragraph{Input projection.}
The input projection consists of a ResMLP with $L=2$, $C_i$ is the input feature dimension, and $C_\text{h} = C_\text{o}$ are set to $C$, the feature dimension of the model.

\paragraph{Output projection.}
The output projection consists of a Layer Norm \citep{ba2016layer} followed by a ResMLP with $C_\text{i}=C$, $L=2$, and $C_\text{o}$ is the output label dimension.

\subsection{FLARE block}
The FLARE block illustrated in \autoref{fig:FLARE}, and detailed in \autoref{sec:method}, consists of the pointwise ResMLP layer, and the FLARE token mixer.
For the ResMLP, we set $C_\text{i} = C_\text{h} = C_\text{o} = C$, set the number of layers to $3$, and allow residuals to flow through the entire block.

\paragraph{FLARE.}
FLARE consists of two ResMLPs for key/value projections, the token operation described in \autoref{fig:FLARE_pseudocode1}, and an output projection.
For the key/value projections, we set $C_\text{i} = C_\text{h} = C_\text{o} = C$, $L=3$, and allow residuals to flow through the entire block.
\autoref{fig:FLARE_pseudocode2} presents a mathematically equivalent PyTorch implementation for multi-head token-mixing operation without the fused SDPA kernels.
The primary memory bottleneck in this implementation is materializing the $M \times N$ encoding weights and the $N \times M$ decoding weights.
Its storage requirement is, thus, $\O(MN)$.
Finally, the output projection is set to a single linear layer.

\section{Spectral analysis}
\label{sec:appendix_spectral}

\begin{figure}[t]
\centering
\begin{lstlisting}[language=Python]
import torch.nn.functional as F
def flare_multihead_mixer_inefficient(Q, K, V):

    # Args - Q: [H M D], K, V: [B H N D]
    # Ret - Y: [B H N D]

    # Compute projection weights
    scores = Q @ K.mT                       # [B H M N]
    W_encode = F.softmax(scores, dim=-1)    # [B H M N]
    W_decode = F.softmax(scores.mT, dim=-1) # [B H N M]

    # Encode: Project to latent sequence (M tokens)
    Z = W_encode @ V

    # Decode: Project back to input space (N tokens)
    Y = W_decode @ Z

    return Y
\end{lstlisting}
\caption{
Pseudocode of FLARE if attention kernel is not available.
See \autoref{fig:FLARE_pseudocode1} for efficient implementation.
}
\label{fig:FLARE_pseudocode2}
\end{figure}

\subsection{Eigenanalysis procedure}
\label{sec:appendix_spectral_method}

We exploit the low-rank structure of the global communication matrix
$W = W_{\text{decode}} \cdot W_{\text{encode}} \in \R^{N \times N}$
to obtain its eigendecomposition in $\O(M^3 + M^2 N)$ time, compared to the $\O(N^3)$ cost for a dense communication matrix.
We find the eigendecomposition of the FLARE attention matrix without actually forming the $N \times N$ matrix.
We first note that $W_{\text{encode}}$ and $W_{\text{decode}}$ can be written in terms of the exponentiated score matrix
$A = \exp(Q \cdot K^T) \in \R^{M \times N}$ as
\eqn{
    W_{\text{encode}} = \Lambda_M \cdot A, \text{ and }
    W_{\text{decode}} = \Lambda_N \cdot A^T,
}
where
$\Lambda^M \in \R^{M \times M}$,
and
$\Lambda^N \in \R^{N \times N}$
are diagonal matrices whose entries are 
\eqn{
    \relax
    [\Lambda_M]_m = \frac{1}{\sum_{n=1}^{N} [A]_{m,n}}, \text{ and }
    [\Lambda_N]_n = \frac{1}{\sum_{m=1}^{M} [A]_{m,n}}.
}
Thus we have
\eqn{
    W =  \Lambda_N A^T \Lambda_M A
}
as the low-rank attention matrix.
We observe that $W$ is similar to $J^T J \in \R^{N \times N}$
where
$J = \Lambda_M^{1/2} A \Lambda_N^{1/2} \in \R^{M \times N}$.
This is because
\eqn{
    W
    &= \Lambda_N A^T \Lambda_M A
    =
    \underbrace{(\Lambda_N^{1/2} \Lambda_N^{1/2})}_{\Lambda_N}
    A^T
    \underbrace{(\Lambda_M^{1/2} \Lambda_M^{1/2})}_{\Lambda_M}
    A
    \underbrace{(\Lambda_N^{-1/2} \Lambda_N^{1/2})}_{I_N}
    \\
    &=
    \Lambda_N^{1/2}
    \underbrace{(\Lambda_N^{1/2} A^T \Lambda_M^{1/2})}_{J^T}
    \underbrace{(\Lambda_M^{1/2} A \Lambda_N^{1/2})}_{J}
    \Lambda_N^{-1/2}.
}
Thus $ J^T J $ is symmetric, positive semi-definite, with rank at most $ M $.
Now suppose a singular value decomposition of $J$ as
$J = U \Sigma V^T$ where $U \in \R^{M \times M}$ and $V \in \R^{N \times M}$ are the matrices whose columns are the left and right singular vectors of $J$ respectively, and $\Sigma \in \R^{M \times M}$ is the diagonal matrix of singular values.
Then, we obtain
\eqn{
    J^T J
    = V \Sigma \underbrace{U^T U}_{I_M} \Sigma V^T
    = V \Sigma^2 V^T,
}
and
\eqn{
    W = \Lambda_N^{1/2} V \Sigma^2 V^T \Lambda_N^{-1/2}.
}
Post-multiplying both sides by $\Lambda_N^{1/2}V$, we have
\eqn{
    W (\Lambda_N^{1/2} V) = (\Lambda_N^{1/2} V) \Sigma^2.
}
Therefore, the $M$ nonzero eigenvalues of $W$ are the squares of the singular values of $J$,
and the corresponding eigenvectors are the columns of $\Lambda_N^{1/2} V$.
Obtaining the eigenvalues and eigenvectors of $W$ this way requires the singular value decomposition of $J\in \R^{M \times N}$.
We can do better by relating $V$ and $\Sigma$ to the eigen decomposition of $JJ^T$.
Consider the matrix $J J^T \in \R^{M \times M}$ with singular value decomposition, we have
\eqn{
    J J^T
    = U \Sigma \underbrace{V^T V}_{I_M} \Sigma U^T
    = U \Sigma^2 U^T.
}
We note that the nonzero eigenvalues of $W$ are the same as the singular values of $JJ^T$.
To obtain the eigenvectors of $W$, we need an expression for $V$ in terms of $U$, $J$, and $\Sigma$.
We do so by noting that
\eqn{
    J^T U = (V \Sigma U^T) U = V \Sigma
    \implies
    V = J^T U \Sigma^{-1}.
}
Therefore, the eigenvectors of $W$ are
\eqn{
    \Lambda_N^{1/2} V
    = \Lambda_N^{1/2} J^T U \Sigma^{-1}.
}

\begin{algorithm}[t]
\label{alg:spectra}
\caption{Eigenvalues and Eigenvectors from $Q, \, K$}
\begin{algorithmic}[1]
    \REQUIRE $Q\in\R^{M \times D}, \, K\in\R^{N \times D}$
    \STATE $A \gets \exp(Q \cdot K^T)$
    \STATE $L_N \gets \text{diag}(1/\sum_{m=1}^M [A]_{m,n})$
    \STATE $L_M \gets \text{diag}(1/\sum_{n=1}^N [A]_{m,n})$
    \STATE $J \gets L_M^{1/2} \cdot A \cdot L_N^{1/2}$
    \STATE Compute SVD: $U \Sigma^2 U^T \gets JJ^T \in \R^{M \times M}$
    \STATE $\text{Eigenvalues} \gets \Sigma^2$
    \STATE $\text{Eigenvectors} \gets L_N^{1/2} J^T U \Sigma^{-1}$
\end{algorithmic}
\end{algorithm}

To find the eigenvalues and eigenvectors of $W$, one only needs to compute the eigen-decomposition of the $M \times M$ matrix $J^T J$.
The overall algorithm, summarized in Algorithm 1, takes
$\O(M^3 + NM^2)$ time where the $\O(M^3)$ is for computing the SVD of $JJ^T$.

\subsection{Qualitative Analysis}
\label{sec:appendix_spectra_qualitative}
The matrix $ J J^T $, being $ M \times M $, captures the structure of this latent space and provides insights into how these $ M $ dimensions contribute to the attention mechanism.
Large eigenvalues correspond to dominant latent dimensions that contribute significantly to attention patterns.
If some eigenvalues are small or zero, those latent dimensions contribute little, suggesting redundancy in the latent space.
The number of nonzero eigenvalues gives the effective rank of $ W $, which reflects how many independent patterns the attention mechanism captures.

\autoref{fig:spectra} (middle) presents the $M=64$ nonzero eigenvalue spectra of an FLARE model trained on the elasticity dataset with $M=64$ latents.
Some observations are as follows.
In all blocks, especially block 1, the eigenvalues drop sharply within the first $10-20$ indices.
This indicates that even though the communication matrices $W_h$ could have rank up to $M=64$, most of the energy (information) is captured by a much smaller subset of modes.
This result validates the hypothesis that the global communication pattern is inherently low-rank.

We also observe that the eigenvalue curves in block 5 and block 8 decay more slowly, retaining more moderate-magnitude eigenvalues beyond index 20–30.
This indicates that as depth increases, the attention mechanism seems to leverage more of the latent space — i.e., the effective rank increases.
This shows that deeper layers learn richer global dependencies, and the model may be using more of the projection capacity in later blocks.

Finally, we note that the curves for different heads (colors) have distinct decay patterns, especially in later blocks.
This reinforces the claim that separate projection matrices per head enables specialized routing.
This supports the idea that FLARE benefits from head-wise diversity, rather than using shared projections like in Transolver \citep{wu2024transolver}.


\section{Benchmarking and Comparison}

\subsection{Benchmark metrics}
The primary evaluation metric for all benchmarks is the relative $\L_2$ error, which quantifies the normalized discrepancy between the predicted solution $\hat{u}$ and the ground truth solution $u$ over all points in the domain. For a given test sample, the relative $L_2$ error is defined as:
\begin{equation}
    \text{Relative } L_2 = \frac{\|\hat{u} - u\|_2}{\|u\|_2}
\end{equation}
where $\|\cdot\|_2$ denotes the standard Euclidean norm.
For datasets where each sample consists of $N$ points (or grid locations), this expands to:
\begin{equation}
    \text{Relative } L_2 = \frac{\left( \sum_{i=1}^{N} (\hat{u}_i - u_i)^2 \right)^{1/2}}{\left( \sum_{i=1}^{N} u_i^2 \right)^{1/2}}.
\end{equation}

The reported metric is averaged over all test samples.

\begin{table}[t]
    \centering
    \caption{
    Summary of PDE benchmarks.
    }
    \label{tab:dataset_summary}
    \begin{tabular}{l|ccccc}
        \toprule
        \textbf{Benchmark} &
        \textbf{\#Dim}     &
        \textbf{Grid Type} &
        \textbf{\#Points}  &
        \makecell{\textbf{\#Input/ Output}\\ \textbf{Features}} &
        \makecell{\textbf{\#Train/ Test}  \\ \textbf{Cases}}    \\

        \midrule
        Elasticity & 2D & Unstructured &    972        & 2 / 1 & 1000 / 200 \\
        \midrule  
        Darcy      & 2D & Structured   &  7,225        & 1 / 1 & 1000 / 200 \\
        Airfoil    & 2D & Structured   & 11,271        & 2 / 1 & 1000 / 200 \\
        Pipe       & 2D & Structured   & 16,641        & 2 / 1 & 1000 / 200 \\
        \midrule
        DrivAerML-40k    & 3D & Unstructured & 40,000        & 3 / 1 &  387 / 97  \\
        LPBF       & 3D & Unstructured & 1,000--50,000 & 3 / 1 & 1100 / 290 \\
        \bottomrule
    \end{tabular}
\end{table}

\subsection{Benchmark datasets}
\label{sec:benchmark_datasets}
We evaluate all models on five benchmark datasets and our proposed AM dataset, each designed to assess generalization, scalability, and robustness to domain irregularity in PDE surrogate modeling. A summary is provided in \autoref{tab:dataset_summary}.

\paragraph{Elasticity.}
This benchmark estimates the inner stress distribution of elastic materials based on their structure. Each sample consists of a tensor of shape $972 \times 2$ representing the 2D coordinates of discretized points, and the output is the stress at each point ($972 \times 1$). The dataset contains 1,000 training and 200 test samples with different material structures~\citep{li2023fourier}.

\paragraph{Darcy.}
This benchmark models fluid flow through a porous medium.
The porous structure is discretized on a $421 \times 421$ regular grid, downsampled to $85 \times 85$ for main experiments.
The output is the fluid pressure at each grid point.
There are 1,000 training and 200 test samples with varying medium structures~\citep{li_fourier_2021}.

\paragraph{Airfoil.}
This task estimates the Mach number distribution around airfoil shapes. The input geometry is discretized into a structured mesh of shape $221 \times 51$, representing deformations of the NACA-0012 profile, and the output is the Mach number at each mesh point. The dataset includes 1,000 training and 200 test samples with unique airfoil designs~\citep{li2023fourier}.

\paragraph{Pipe.}
This benchmark predicts horizontal fluid velocity in pipes with varying internal geometries. Each sample is represented as a $129 \times 129 \times 2$ tensor encoding mesh point positions, with the output being the velocity at each mesh point ($129 \times 129 \times 1$). The dataset consists of 1,000 training and 200 test samples, with pipe shapes generated by controlling the centerline~\citep{li2023fourier}.


\paragraph{DrivAerML-40k.}
The DrivAerML dataset~\citep{ashton2024drivaerml} has high-fidelity automotive aerodynamic simulations, featuring 500 parametrically morphed DrivAer vehicles.
CFD simulations are performed on 160 million volumetric mesh grids using hybrid RANS-LES, the highest-fidelity scale-resolving CFD approach routinely deployed by the automotive industry~\citep{ashton2024drivaerml}.
Each sample in the dataset includes a surface mesh with approximately 8.8 million points and corresponding pressure values.
Since no official dataset split is provided, an 80/20 random split is used, with 40,000 points sampled per case for training and evaluation.

\paragraph{LPBF.}
We introduce the laser powder bed fusion (LPBF) additive manufacturing dataset which involves field prediction on complex geometries.
In metal additive manufacturing, variations in design geometry can affect the dimensional accuracy of the part and lead to shape distortions.
We perform several LPBF process simulations of a set of geometries to obtain the deformation field over the geometry.
We select a subsample of the dataset with up to 50,000 points per geometry and divide it into 1,100 training cases and 290 test cases.
We train models to learn the $Z$ (vertical) component of the deformation field at each grid point.
Additional details are provided in \autoref{sec:appendix_am_dataset}.

\subsection{Benchmark models and training details}
\label{sec:appendix_benchmark_models}

\begin{table}
\centering
\caption{Standard training configuration on PDE datasets. Identical values are grouped for clarity.}
\label{tab:training_hyperparams}
\begin{tabular}{lccccc}
\toprule
\textbf{Dataset} & \textbf{Batch Size} & \textbf{Weight Decay} & \textbf{Learning Rate} & \textbf{Epochs} & Loss \\
\midrule
Elasticity    & 2 & $10^{-5}$  & $10^{-3}$ & 500 & Rel. $L_2$ \\
\midrule
Darcy         & 2 & $10^{-5}$  & $10^{-3}$ & 500 & Rel. $L_2$ + 0.1 $L_\text{g}$ \\
Airfoil       & 2 & $10^{-5}$  & $10^{-3}$ & 500 & Rel. $L_2$ \\
Pipe          & 2 & $10^{-5}$  & $10^{-3}$ & 500 & Rel. $L_2$ \\
\midrule
DrivAerML     & 1 & $10^{-2}$  & $10^{-3}$ & 500 & Rel. $L_2$ \\
LPBF          & 1 & $10^{-4}$  & $10^{-3}$ & 250 & Rel. $L_2$ \\
\bottomrule
\end{tabular}
\begin{tablenotes}
\item Note: For Darcy test case, we include an additional gradient regularization term $L_g$ following Transolver \citep{wu2024transolver}.
\end{tablenotes}
\end{table}

\begin{table}
\centering
\caption{Model configurations for FLARE for PDE datasets. Identical values are grouped for clarity.}
\label{tab:model_config_FLARE}
\begin{tabular}{lcccc}
\toprule
\textbf{Dataset} & \textbf{\#Heads  ($H$)} & \textbf{\#Latents ($M$)} & \textbf{\#Blocks ($B$)} & \textbf{\#Features $(C)$} \\
\midrule
Elasticity    & 8  & 64  & 8  & 64  \\
\midrule
Darcy         & 16 & 256 & 8  & 64  \\
Airfoil       & 8  & 256 & 8  & 64  \\
Pipe          & 8  & 128 & 8  & 64  \\
\midrule
DrivAerML-40k & 8  & 256 & 8  & 64  \\
LPBF          & 16 & 256 & 8  & 64  \\
\bottomrule
\end{tabular}
\end{table}

We follow the recommended hyperparameter configuration for LNO, Transolver, and GNOT wherever possible.
For consistency, we set the number of blocks to $B=8$,
and strive to match the parameter counts of Transolver without (w/o) convolution for all models.
As such, the hidden feature dimension for vanilla transformer, is set to $C=80$.
Its head dimension is set to $D=16$ and MLP ratio is 4 which is typical for transformers \citep{vaswani2017attention}.
For FLARE, we set the hidden feature dimension to $C=64$, and use $H=8$ or $16$ heads, which result in a head dimension of $D=8$ or $4$ respectively.
The number of latents is chosen from $M\in\{64,\, 128,\, 256\}$ depending on the problem.
As PerceiverIO was not designed to be a surrogate model,
we generously set its channel dimension to $C=128$, and number of latents to $M=1,024$.
Furthermore, the input and output projections for vanilla transformer, perceiver, and FLARE are held consistent to facilitate an equitable comparison of their point-to-point communication schemes.

We evaluate Transolver, GNOT with the hyperparameter configurations provided in \cite{wu2024transolver} and LNO on the ones provided in \cite{wang2024latent}
on the 2D test cases.
For the remaining problems, we choose the best performing parameter set from the 2D cases.

Unless otherwise stated, all models are trained with the AdamW \citep{loshchilov_decoupled_2019} optimizer ($\beta_1 = 0.9$, $\beta_2 = 0.999$) to minimize the relative $L_2$ error.
We use the OneCycleLR \citep{smith2019super} scheduler with 10\% of epochs dedicated to warming-up to a learning rate of $10^{-3}$, followed by cosine decay.
We train on LPBF for 250 epochs, and 500 epochs for all other models.
Note that we use gradient clipping with max\_norm$=1.0$ unless otherwise stated.
Unless otherwise stated, the weight decay regularization parameter is set to $10^{-5}$ for the 2D test cases,
$10^{-4}$ for DrivAerML-40k,
and $10^{-4}$ for LPBF.
The batch size is set to 2 for the 2D problems and 1 for the 3D problems unless otherwise stated.
Unless otherwise stated, all models are trained in full precision (FP32).

\paragraph{Vanilla transformer.}
For the vanilla transformer, we set the hidden feature dimension to $C=80$, and choose $H=5$ heads so that the head dimension $D=16$.
The number of blocks is set to $B=8$, and the MLP ratio for the feedforward block is set to 4.
The vanilla transformer can be prohibitively slow for test cases with over 10,000 test points.

\paragraph{PerceiverIO.}
For PerceiverIO, we use $B=8$, $C=128$, $H=8$, and set the latent sequence length to $M=512$ for all test cases.

\paragraph{Transolver.}
Transolver (\cite{wu2024transolver}) introduces Physics-Attention: each mesh point is softly assigned to a few learnable \emph{slices}, shrinking thousands of points to only tens of tokens.
Self-attention runs on this compact set and the tokens are then desliced back to points.
Following the hyperparameter recommendations in the code of \cite{wu2024transolver},
Transolver is trained with 30\% of the steps dedicated to warm-up, and gradient clipping with max\_norm $=0.1$.
The recommended batch size for Transolver is 1 for elasticity, and 4 for the remaining 2D problems.
For the 3D problems, we set the batch size to 1.

\paragraph{Transolver++.}
We evaluated Transolver++ using the latest version of the official code release (\url{https://github.com/thuml/Transolver_plus}) together with the hyperparameters reported in the paper.
All experiments were run using PyTorch 2.7 and CUDA 12.8; the supplementary material provides a script to instantiate the standard environment used for all benchmark runs.

\paragraph{Latent Neural Operator (LNO).}
LNO (\cite{wang_latent_2024}) moves computation into a small latent space.
An embedder lifts the input field, cross-attention compresses $N$ points into $M$ latent tokens, transformers act solely on these tokens, and a decoder maps them to any query location.
In line with the recommended hyperparameter configuration in \cite{wang_latent_2024},
we set $\beta_2=0.99$ in the AdamW optimizer,
do warmup for the first 20\% of epochs,
and clip gradient norms greater than 1,000.
The number of hidden features are set to $C=192$ for elasticity, and $C=128$ for all other test cases.
The number of residual layers is set to 3 for elasticity and 4 for other test cases.
The number of latent self attention blocks is 8 for pipe and airfoil test cases and 4 for all other cases.
The number of latent modes is set to 256 for all test cases.
The batch size during training is set to 4 for the 2D test cases and 1 for the 3D test cases.
The LNO code also recommends weight decay regularization of $5\cdot10^{-5}$ for the 2D test cases.

Through communication with the LNO authors (see \url{https://github.com/L-I-M-I-T/LatentNeuralOperator/pull/6}),
we learned that the datasets used in the LNO paper and in the original Transolver paper are not the same.

In running our experiments, we noticed discrepancies between our LNO results and those presented in their article \citep{wang2024latent}.
Upon further investigation, we found that the datasets used in the LNO paper and in the original Transolver paper are not the same.
For example, in the Elasticity dataset, LNO was trained and tested on a $1000/1000$ split, whereas Transolver used a $1000/200$ split.
In the Darcy dataset, LNO employed a higher resolution of $241 \times 241$, compared to $85 \times 85$ in Transolver.
Because these differences make direct comparison unreliable, all models (including LNO and Transolver) were re-trained and evaluated on the standardized Transolver splits and resolutions to ensure fairness.


\paragraph{General Neural Operator Transformer (GNOT).}
GNOT (\cite{hao2023gnot}) employs heterogeneous normalized attention, separately normalizing keys and values, to fuse multiple input fields on irregular meshes. A learnable geometric gate decomposes the domain and routes tokens to scale-specific expert MLPs. Linear cost attention plus this gating scales to large problems and surpasses earlier operator learners.

Following the hyperparameter recommendations outlined in the code of \cite{wu2024transolver}, 
GNOT is trained with 30\% of the steps dedicated to warm-up, and gradient clipping with max\_norm = $0.1$.
The recommended batch size for Transolver is 1 for elasticity, and 4 for the remaining 2D problems.
The batch size is set to 2 for elasticity, 4 for the remaining 2D benchmarks and 1 for the 3D benchmarks.

\paragraph{FLARE.}
All attention operations in FLARE are implemented using \texttt{scaled\_dot\_product\_attention} (SDPA) in \texttt{torch.nn.functional}, as shown in \autoref{fig:FLARE_pseudocode1}; this allows PyTorch to dispatch to fused SDPA backends (e.g., FlashAttention-style kernels) when available.
For all problems, we employ $B=8$ blocks with a feature dimension of $C=64$.
We set the number of residual layers and the number of key/value projection layers to 3, and vary the head dimension as
$D\in\{4,8\}$,
the number of latent tokens as
$M\in\{64, 128, 256\}$
The hyperparameters for each test case are presented in \autoref{tab:model_config_FLARE}.

\begin{figure}[t]
    \centering
    \includegraphics[width=1.0\linewidth]{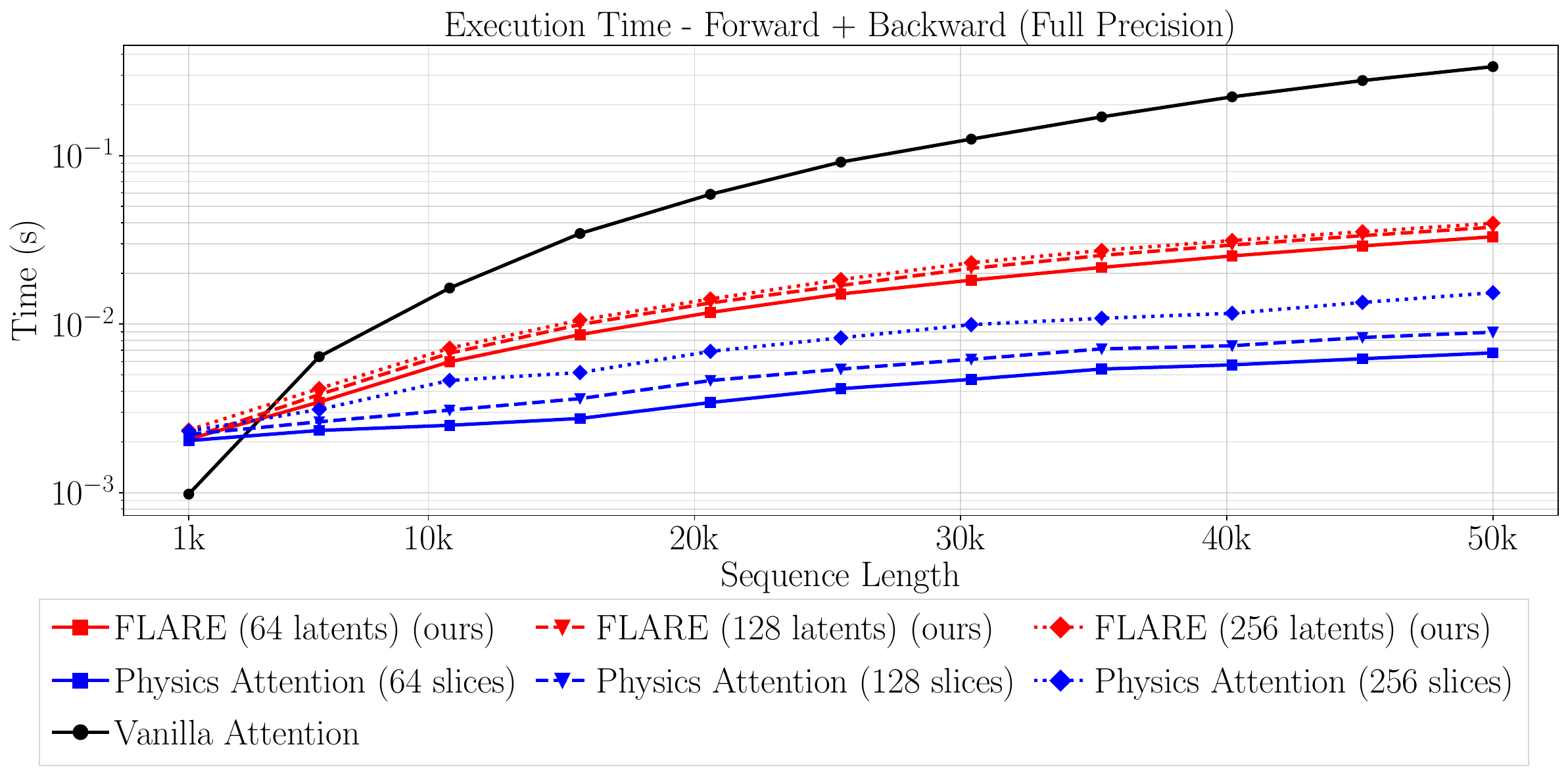}
    \caption{
    Execution times in FP32 for a single vanilla self-attention layer, a physics attention layer, and FLARE.
    The models are set to have approximately the same number of parameters as in \autoref{sec:experiments_pde_surrogates}.
    This calculation is performed on a single H100 80GB GPU.
    Note that the curves for FLARE are somewhat overlapping.
    }
    \label{fig:time_memory_fp32}
\end{figure}

\section{Field-prediction on million-point geometries}
\label{sec:appendix_field_pred}

\paragraph{Experimental setup.}
We train FLARE on the DrivAerML dataset \citep{ashton2024drivaerml} using meshes subsampled to contain $N=10^6$ points per geometry.
All experiments are run on a single Nvidia H100 80GB GPU with batch size $1$.
Unless otherwise stated, models use feature dimension $C=64$ and $H=8$ attention heads.
We vary the number of latent tokens $M\in\{128,512,1024,2048\}$ and the number of FLARE blocks $B$ (as reported in \autoref{fig:scale_dml}).

\paragraph{Training details.}
All FLARE models in this study are trained in mixed precision to take advantage of fused SDPA backends (e.g., FlashAttention-style kernels).
We train for 500 epochs with the OneCycleLR scheduler \citep{smith2019super} where the first $5\%$ of epochs are spent warming up to a learning rate of $5\cdot10^{-4}$ followed by cosine decay.
No hidden sparsity, memory offloading, or distributed training is used.

\section{Discussion on the design principles of FLARE.}
\label{sec:appendix_flare_design}

\paragraph{Latent tokens enable gather-scatter communication.}
In FLARE, information flows through latent tokens by first \textit{gathering} from the input sequence and then \textit{scattering} back.
The encoding step can be understood as a gather (all-reduce) operation, where each latent token pools information from the input according to its learned query pattern.
Formally, for latent query $q_m$,
\begin{equation}
z_m = \sum_{n=1}^N 
\frac{\exp(q_m \cdot k_n)}{\sum_{n'=1}^N \exp(q_m \cdot k_{n'})} \, v_n,
\quad m = 1, \dots, M,
\end{equation}
$z_m$ aggregates input values $v_n$ with convex weights.
When the similarity scores are sharp, $z_m$ emphasizes a few dominant inputs; when they are flatter, $z_m$ acts like a specialized \textit{pooling token} that averages a select set of tokens.
Across $M$ latents, this yields a compact set of global descriptors, each specializing in pooling different aspects of the input.

The decoding step is the dual scatter (broadcast) operation, but importantly, the broadcast is \textit{selective}: each latent $z_m$ contributes only to the input tokens that assigned it high weight in the encoding step.
Concretely,
\begin{equation}
y_n = \sum_{m=1}^M 
\frac{\exp(k_n \cdot q_m)}{\sum_{m'=1}^M \exp(k_n \cdot q_{m'})} \, z_m,
\quad n = 1, \dots, N,
\end{equation}
the output token $y_n$ only receives substantial information from the latents whose query pattern matches its key $k_n$ strongly.
In this sense, each latent acts as both a selective \textit{pooling hub} and \textit{broadcaster}, routing information back along the pathways that originally recruited it.
Together, these gather-scatter stages form a low-rank, butterfly-like transformation: first contracting $N$ tokens into $M \ll N$ global features, and then expanding back to $N$ outputs, enabling efficient, yet expressive, global communication.

\paragraph{Symmetry in latent token communication.}
The compression–expansion attention structure in FLARE is most effective when the encoding and decoding operators are structurally aligned.
Specifically, $W_{\text{encode}, h} = \softmax(Q_h \cdot K_h^T)$ and $W_{\text{decode}, h} = \softmax(K_h \cdot Q_h^T)$ are transposes of each other up to diagonal scaling.
We experimented with breaking this symmetry by using distinct query–key pairs for encoding and decoding, but observed no accuracy gains.
This suggests that the near-adjoint relationship between $W_{\text{encode}}$ and $W_{\text{decode}}$, both derived from the same parameters, may provide a form of mathematical optimality, ensuring stable information flow through latent tokens while reducing representational redundancy.

\paragraph{Tradeoff between query dynamics and key/value expressivity.}
An important design choice in FLARE is to use fixed, input-independent queries $Q$, which constrain the flexibility of the attention pattern.
Detaching $Q$ from $X$ allows for the low-rank communication structure, but requires compensating expressivity in the key and value projections.
In practice, we find that deeper residual MLPs for $K$ and $V$ are crucial for capturing rich, feature interactions under this constraint.
Conversely, one could imagine making $Q$ dynamic by conditioning it on $X$, which would shift more of the modeling burden to the query side and potentially allow shallower $K, V$ projections.
Thus, FLARE embodies a clear tradeoff: fixing $Q$ encourages stability and efficiency, but places greater importance on the depth and expressivity of the key/value encoders.

\section{Model analysis and ablations}
\label{sec:appendix_ablation}

\begin{figure}[t]
    \centering
    \includegraphics[width=0.494\linewidth]{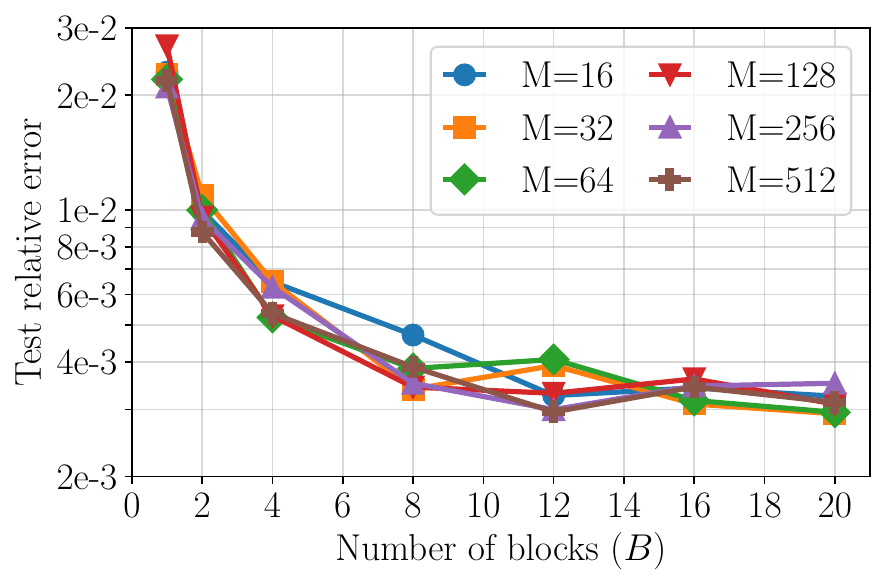}
    \includegraphics[width=0.494\linewidth]{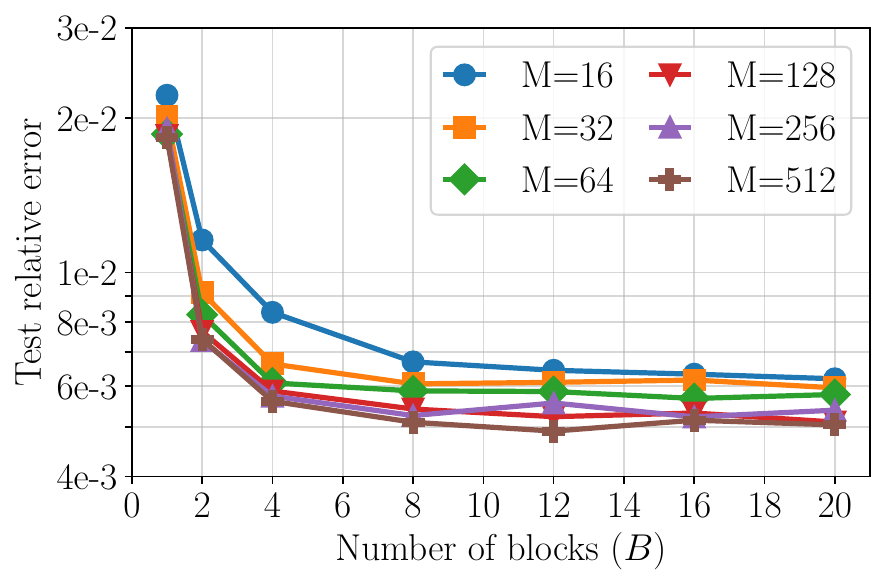}
    \caption{
    Effect of number of blocks ($B$) and number of latent tokens ($M$) on test accuracy on the Elasticity (left) and Darcy (right) test cases. Experiment details are presented in \autoref{sec:appendix_ablation}.
    }
    \label{fig:num_blocks}
\end{figure}

\paragraph{Time and memory complexity.}
\autoref{fig:time_memory} illustrates the time and memory complexity of a single forward and backward pass for different attention schemes on long sequences.
The experiment is done in mixed precision (FP16 in forward pass, FP32 in backward pass) using PyTorch's autocast functionality with $C=128$ features and $H=8$ heads for all models.
The FlashAttention backend \citep{dao2022flashattention} is employed for SDPA wherever possible.

Although vanilla self-attention has the lowest memory cost thanks to the FlashAttention algorithm, which eliminates the need to materialize the score matrices ($Q_h \cdot K_h^T$), its compute time still scales poorly with the sequence length.
In contrast, the compute time for FLARE exhibits strong scaling with sequence length.
Its memory requirement is marginally greater than vanilla attention due to the presence of deep residual networks for key/value projections, and due to the need to materialize $Z_h$, the latent sequence of $M$ tokens.
As these costs are marginal compared to the SDPA operation, the curves for different $M$ values of FLARE are somewhat overlapping.
Finally, the compute time for Physics Attention of Transolver \citep{wu2024transolver} exhibits somewhat good scaling.
However, its memory cost and compute time blow up for large slice counts due to the need for materializing the projection matrices.

\paragraph{Number of blocks ($B$) and latent tokens ($M$).}
\autoref{fig:num_blocks} presents the test relative error of FLARE on the Elasticity (left) and Darcy (right) benchmark datasets as a function of the number of blocks $(B)$ and the number of latent tokens $(M)$.
\autoref{fig:scale_dml} (left) presents the same for the DrivAerML dataset with one million points per geometry.
In all cases, we note the favorable trend that relative error consistently decreases as we increase the number of blocks.
Similarly, we observe that the relative error generally decreases with $M$, though the trend is not strictly monotonic.
In the Elasticity problem, improvements with rank diminish rapidly, indicating that global communication in that problem is fundamentally low-rank.
On the other hand, increasing $M$ monotonically increases performance on the Darcy problem, indicating that the problem is \emph{rank-limited}.
This also explains why vanilla transformer with a full-rank attention pattern outperforms rank-deficient FLARE on the darcy problem.
However, the accuracy gain comes at the cost of greater latency as the vanilla transformer is $\sim5\times$ slower than FLARE on the Darcy problem.
\autoref{fig:scale_dml} indicates that time per epoch (middle) and memory (right) scaling of FLARE with $B$ and $M$.
Here, increasing $M$ leads to increased latency, and that increasing $M$ does not come at the cost of greater memory requirements.

\begin{figure}[t]
    \centering
    \includegraphics[width=1.0\linewidth]{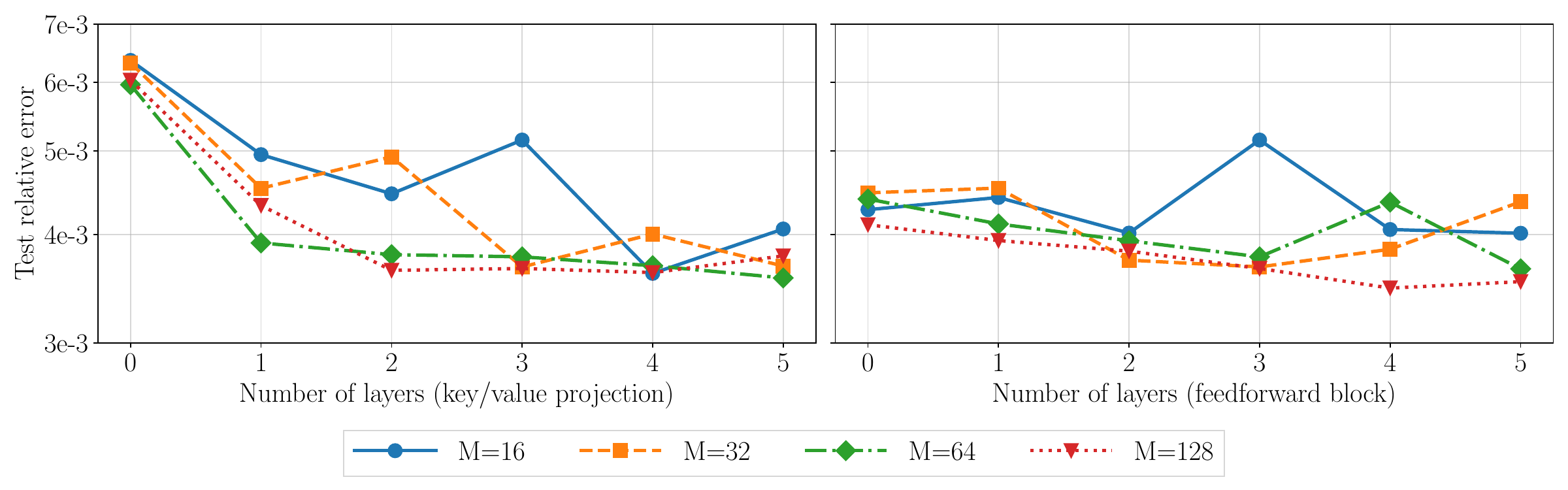}
    \caption{
    (Left) effect of the number of residual layers in key/ value projection, and (right) of residual layers in residual block on test accuracy.
    In both cases, deeper networks lead to greater accuracy.
    }
    \label{fig:num_layers}
\end{figure}

\begin{figure}[t]
    \centering
    \includegraphics[width=0.7\linewidth]{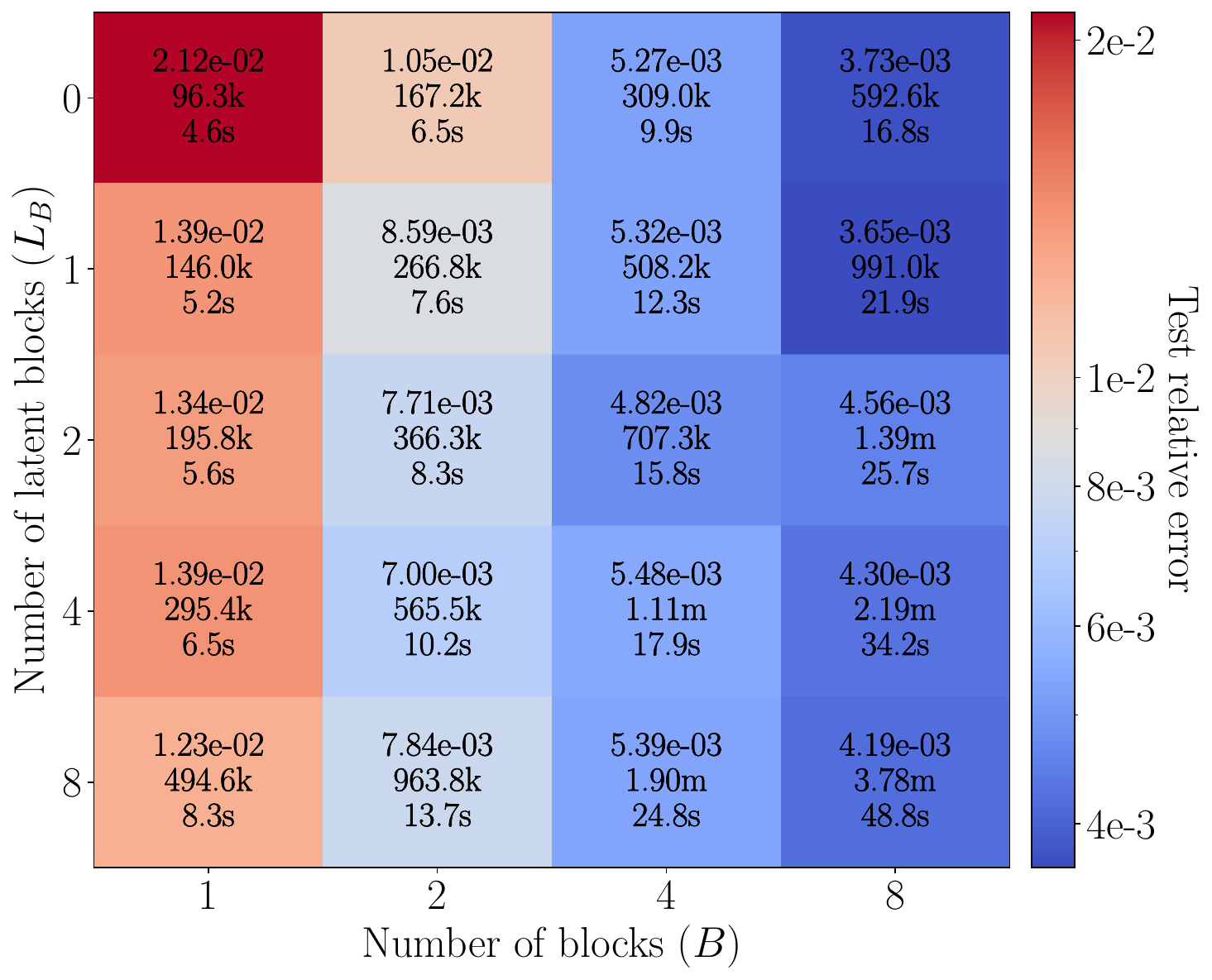}
    \caption{
    \textbf{Ablation on the number of latent-space self-attention blocks ($L_B$) versus the number of FLARE encode-decode blocks ($B$).}
    Each cell reports: (top) test relative error, (middle) parameter count, and (bottom) training time per epoch.
    The total cost of a model with $B$ FLARE blocks and $L_B$ latent blocks is 
    $\mathcal{O}(B\,N M + B L_B\,M^2)$.
    The results show that increasing the number of latent-space blocks, as done in LNO-style and Perceiver-style models, yields worse accuracy and poorer speed/parameter tradeoffs than allocating compute to additional FLARE blocks.
    The optimal regime lies near the \emph{top-right corner}: many encode-decode blocks and \emph{zero} latent-space blocks.
    }
    \label{fig:latent_ablation}
    \vspace{-1\baselineskip}
\end{figure}

\begin{figure}[t]
    \centering
    \includegraphics[width=0.8\linewidth]{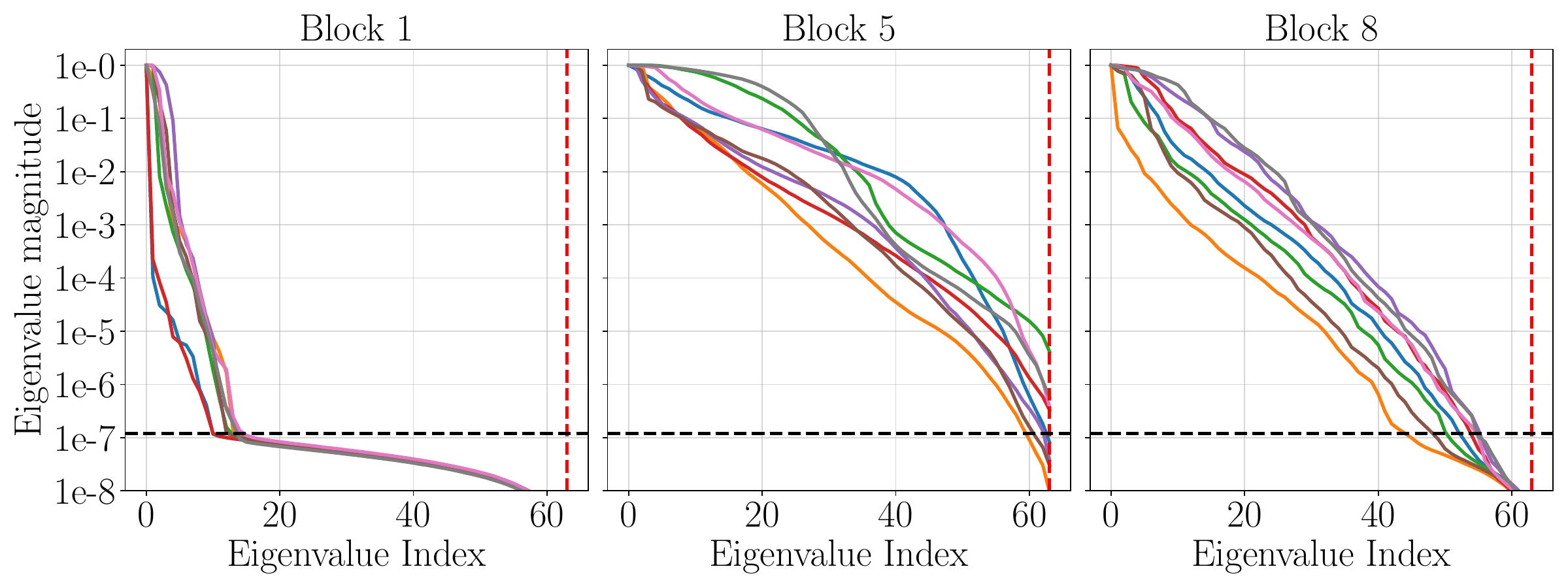}
    \includegraphics[width=0.8\linewidth]{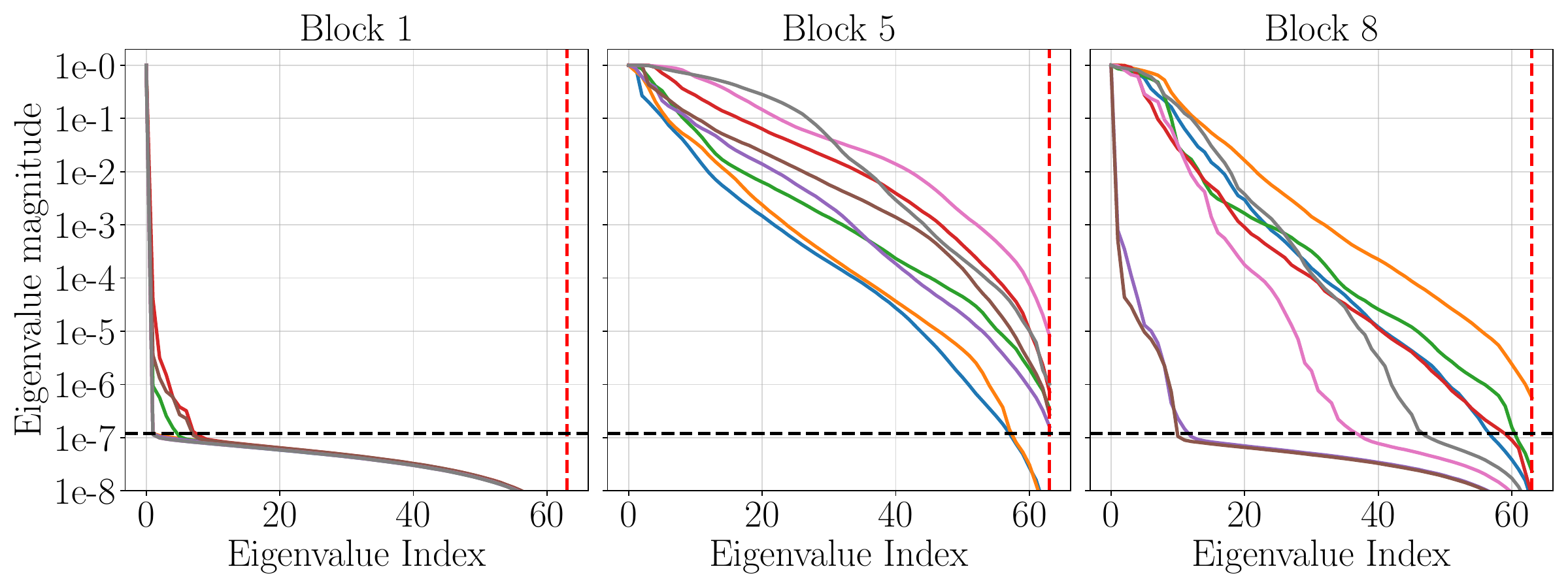}
    \begin{tabular}{r|c|c}
        \toprule
         {Blocks} & {Shared latents} & {Independent latents} \\
         \midrule
         1 & \textbf{20.8} & 21.9 \\
         2 & 10.8 & \textbf{9.99} \\
         4 & 5.91 & \textbf{5.21} \\
         8 & 4.37 & \textbf{3.85} \\
         \bottomrule
    \end{tabular}
    \caption{
    \textbf{Ablation of shared vs.\ independent latent tokens across attention heads.}
    Each plot shows the $M{=}64$ nonzero eigenvalues of the head-specific communication matrices $W_h$ (\autoref{eq:attn_W}) for FLARE with $B{=}8$ blocks and $C{=}64$ features trained on the elasticity dataset.
    When heads share a single latent sequence (top row), the eigenvalue spectra across heads are nearly identical, indicating similar learned low-rank subspaces.
    When heads use independent latent slices (bottom row), the eigenvalue decay varies noticeably across heads, reflecting more diverse low-rank structures.
    The accompanying table reports test relative errors for different depths $B$, showing that models with independent latents consistently achieve lower error.
    }
    \label{fig:spectra}
\end{figure}

\paragraph{ResMLP depth: key/value projections.}
A substantial distinction between vanilla self-attention and FLARE is the introduction of deep residual blocks for key and value projections in place of simple linear layers.
In standard self-attention, queries ($Q_h$) and keys ($K_h$) determine the global communication pattern through $W_h = \softmax(Q_h K_h^T / s)$, while values ($V_h$) carry the information to be communicated.
All three are typically computed as shallow linear projections of the input $X$.
In contrast, FLARE computes the attention pattern as $W_h = \softmax(K_h Q_h^T) \cdot \softmax(Q_h K_h^T)$, where the query embeddings $Q_h$ are learned parameters independent of the input.
This makes the attention pattern less dynamic, motivating architectural modifications to enhance flexibility.

To address this, we replace the linear key and value projections with deep residual MLPs.
Using residual networks for key and value encodings allows each token to learn richer and more structured features rather than shallow embeddings, which is particularly crucial in FLARE since the queries are fixed and cannot adapt to the input.
\autoref{fig:num_layers} (left) shows the impact of varying the number of residual layers in key/value projections and within the residual block on test accuracy for the elasticity benchmark dataset.
We suspect that deeper key/value encodings lead to more meaningful and focused attention, encoding structured inductive priors beneficial to downstream prediction.

\paragraph{ResMLP depth: feedforward block.}
A second difference between the standard attention block and FLARE block is that we replace the feed-forward block in vanilla self-attention with a deep residual MLP.
Preliminary experiments using standard feed-forward blocks led to training instabilities and poor convergence.
In contrast, residual MLPs consistently enabled stable training and allowed us to increase model capacity.
\autoref{fig:num_layers} (right) indicates that increasing the number of residual layers leads to slight improvements in accuracy.
Based on these results, we use three residual layers in both key/value projections and the residual block, as this provides a good trade-off between model capacity and computational cost in all subsequent experiments.

\begin{wrapfigure}[20]{R}{0.50\textwidth}
    \vspace{-1.5em}
    \centering
    \includegraphics[clip=true,width=1.00\linewidth]{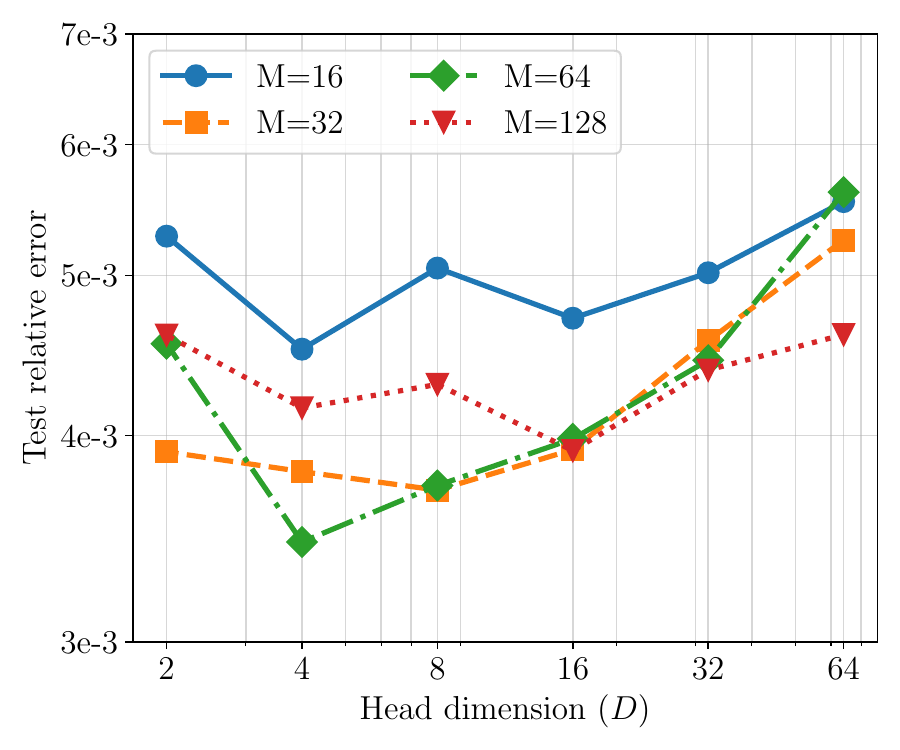}
    \vspace{-1.0em}
    \caption{
    Effect of head dimension ($D$) on test accuracy.
    We design FLARE to work optimally for $D=4$--$8$.
    }
    \label{fig:num_heads}
\end{wrapfigure}

\paragraph{Effect of head dimension and parallel low-rank projections.}
A core hypothesis underlying FLARE is that each attention head implements an independent rank–$\le M$ projection–reconstruction pathway.
When multiple such pathways operate in parallel, the resulting attention operator becomes a mixture of low-rank factors, each capturing a distinct structural component of the underlying communication pattern.
This is in contrast to Transolver, which shares projection weights across all heads, and LNO, which uses a single global projection: both designs restrict the model to learning at most one or a few shared low-rank directions.
FLARE, by contrast, allows each head to specialize in complementary routing patterns by
assigning it an independent slice of the latent tokens.

To examine this hypothesis quantitatively, we vary the number of heads $H$ while keeping the total feature dimension $C=64$ and number of blocks $B=8$ fixed, thereby trading off head dimension $D = C/H$ against the number of parallel low-rank projections available to the model.
As shown in \autoref{fig:num_heads}, FLARE consistently achieves the best accuracy for $D=4$ or $D=8$, outperforming configurations with larger head dimensions.
This behavior is the reverse of what is commonly observed in standard transformers—where typical head
dimensions are $D=16$–$32$, but is entirely consistent with FLARE’s architectural role for each head.
Larger $D$ increases the per-head representational capacity but reduces the number of parallel low-rank factors; smaller $D$ yields more distinct heads, and thereby more distinct projection–reconstruction pathways, which better approximate a full attention map.

We additionally note a practical implication of this analysis: because the dot-product magnitudes are naturally small for $D\in\{4,8\}$, we use a scaling factor of $1$ rather than the usual $1/\sqrt{D}$ in these cases, simplifying implementation without sacrificing stability.
Overall, the head-dimension ablation strongly supports our main architectural claim: FLARE benefits from \emph{multiple parallel low-rank projections}, and enabling per-head independence is crucial for capturing diverse communication patterns.
\paragraph{Ablation on latent-space blocks vs.\ FLARE blocks.}
To directly investigate whether FLARE’s encode-decode mechanism is responsible for the observed performance gains, we conduct a controlled ablation varying the number of latent-space self-attention blocks ($L_B$) and the number of full FLARE blocks ($B$).
This experiment spans the continuum between a Perceiver/LNO-like architecture (few encode-decode operations and many latent-space blocks; bottom-left of \autoref{fig:latent_ablation}) and a FLARE-like architecture (many encode-decode blocks and no latent-space self-attention; top-right).

Across all parameter budgets, we observe a consistent trend: \emph{increasing the number of latent-space blocks degrades accuracy while increasing computational cost}.  
Latent-space self-attention contributes additional parameters and a $\mathcal{O}(M^2)$ cost per block, yet does not improve the model’s ability to capture global structure.  
In contrast, allocating the same compute to additional encode-decode blocks, which perform low-rank global mixing via cross-attention, steadily improves accuracy.  
The lowest errors in the entire grid occur in the top-right corner, corresponding to \emph{zero} latent-space blocks and the largest number of FLARE blocks.

This ablation provides direct causal evidence supporting FLARE’s architectural choice: global communication is most effectively learned by repeating the low-rank projection/unprojection pathway, not by performing nonlinear refinements inside the latent space.  
If one interprets the Perceiver $\leftrightarrow$ FLARE spectrum as trading latent refinement for repeated global mixing, our results show that the optimal regime is decisively FLARE-like.  
For a fixed parameter budget and wall-clock time, the best strategy is to \emph{reduce or eliminate latent-space attention and increase the number of encode-decode blocks}, validating the design principle that token mixing should occur through repeated low-rank attention projections rather than deeper latent-space transformers.

\paragraph{Ablation on shared vs.\ independent per-head latent tokens.}
To assess the effect of head-wise latent diversity—central to FLARE’s design, we compare two variants: (i) \emph{shared-latent} models, where all heads use the same latent sequence, and (ii) \emph{independent-latent} models, where each head receives its own slice of the latent tokens.  
This isolates whether FLARE’s expressivity arises merely from the low-rank encode-decode structure or whether the ability of different heads to learn distinct low-rank projections provides additional modeling capacity.

The spectral plots in \autoref{fig:spectra} reveal a clear difference between the two settings.  
With shared latents, all heads exhibit nearly identical eigenvalue decay, implying that they compress and propagate information in similar ways.  
In contrast, independent latents produce noticeably different spectra across heads, particularly in deeper blocks, indicating that different heads discover complementary low-rank subspaces of the token-to-token communication structure.  
This diversity is modest in early layers but becomes increasingly pronounced as depth increases.

Crucially, the quantitative results mirror this qualitative behavior: across all depths, models with independent per-head latents achieve lower test relative error despite having comparable parameter counts.  
This provides causal evidence—not just descriptive visualization—that head-wise independence is beneficial.  
It supports our claim that FLARE’s mixture of head-specific low-rank projections yields richer communication pathways than architectures that use a single shared projection (e.g., Transolver) or a single latent transformer (e.g., LNO, PerceiverIO).  
Finally, we note that applying the same spectral analysis to Transolver or LNO is not directly meaningful, since their latent-space transformations are nonlinear and not expressible as a single linear low-rank operator; FLARE’s structure uniquely enables such eigenanalysis.

\section{Benchmark dataset of additive manufacturing simulations}
\label{sec:appendix_am_dataset}

\begin{figure}[t]
    \centering
    \includegraphics[width=0.8\linewidth]{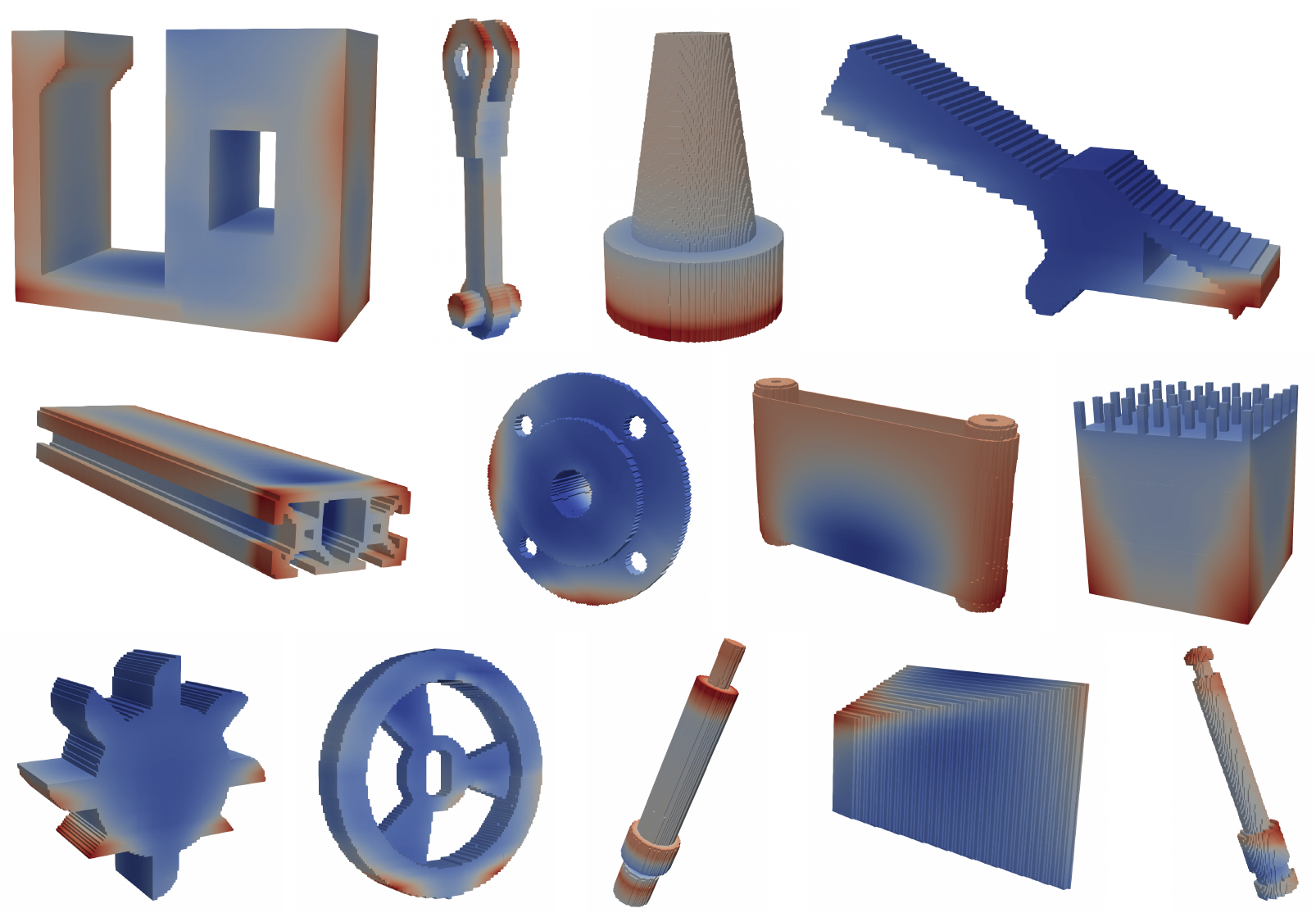}
    \caption{
    We simulate the LPBF process on selected geometries from the Autodesk segmentation dataset \citep{lambourne2021brepnet} to generate a benchmark dataset for AM calculations.
    Several geometries are presented in this gallery.
    The color indicates $Z$ (vertical) displacement field.
    }
    \label{fig:lpbf_gallery}
\end{figure}

\subsection{Introduction \& Background}
In metal additive manufacturing (AM), subtle variations in design geometry and process parameter selection may result in undesirable part artifacts or even costly build failures.
Numerical simulations of laser powder bed fusion (LPBF), a popular additive manufacturing process, can be used to predict build failures, but this may take several minutes to hours depending on the part size.

Although AM enables the fabrication of a wide variety of new designs, the final product must nonetheless comply with the constraints of the underlying manufacturing process.
Metal AM parts often have anisotropic and spatially varying material properties that depend on geometric features (such as overhang, or support structure) and fabrication process parameters (such as laser energy density, hatch spacing, layer thickness, and raster path).
As-built parts with thin features often suffer distortions when manufactured with the LPBF process, which uses a high-power laser to selectively melt and fuse metal powder, layer-by-layer, to create complex, high-precision parts.
Thermal stresses, termed \emph{residual stresses} (RS), accumulate in LPBF-fabricated parts as a result of rapid thermal cycling due to laser exposure.
These stresses can be severe to the point of inducing localized plastic deformations or delamination.
As such, due to these \emph{residual deformations}, the final shape of the part may deviate from the designed geometry.

We present a high-fidelity thermomechanical RS calculation data set on the Fusion 360 segmentation dataset, a publicly available dataset of complex 3D geometries \citep{lambourne2021brepnet}.
Numerical solvers for simulating the LPBF build process perform expensive quasi-static thermo-mechanical equations,
\begin{equation}
    \label{eqn:thermo-mech}
    \underbrace{
    \rho C_p \frac{\d T}{\d t} = \grad \cdot k \Delta T(\boldsymbol{x}, t) + Q(\boldsymbol{x}, t),
    }_{
        \text{thermal transport}
    }
    \hspace{2em}
    \underbrace{
        \grad \cdot \boldsymbol{\sigma} = 0
        \textcolor{white}{\frac{1}{2}}
        \hspace{1em}
        \boldsymbol{\sigma} = \boldsymbol{C} \boldsymbol{\varepsilon}_e,
    }_{
    \text{stress equilibrium}
    }
\end{equation}
layer-by-layer within a finite element framework
\citep{thermo-mechanical,liang_modified_2019,autodesk_help_nodate}, and are integrated in commercial software products \citep{autodesk_help_nodate}.
These calculations take several minutes to hours, making them prohibitively expensive for part design scenarios that can involve hundreds of evaluations.

\citet{ferguson2025topology} introduced a dataset of LPBF simulations in which multiple finite element calculations were performed on a collection of 3D shapes. 
That dataset, however, was restricted to relatively coarse meshes with approximately 3{,}500 points per mesh. 
The present work extends this line of research by considering larger and more refined meshes, with up to 50{,}000 grid points.

\begin{figure}[t]
    \centering
    \includegraphics[width=0.8\linewidth]{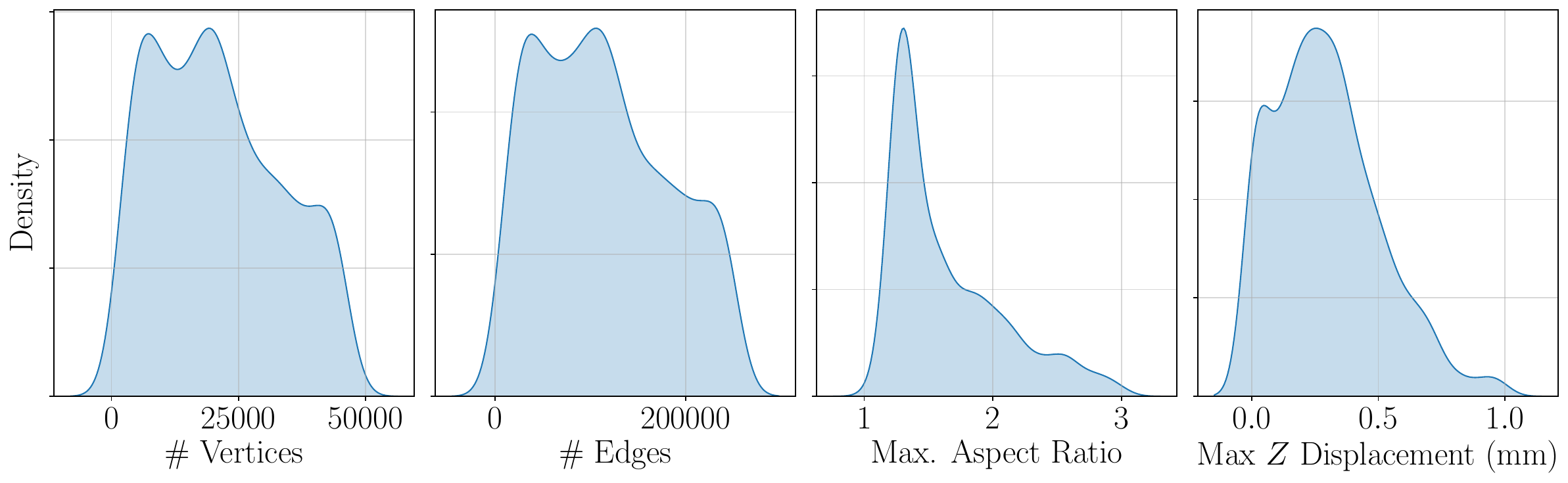}
    \caption{
    Summary of LPBF dataset statistics.
    }
    \label{fig:lpbf_stats}
\end{figure}

\begin{table}[t]
\centering
\caption{Summary statistics of our proposed LPBF dataset.}
\begin{tabular}{l|cccccc}
\toprule
& \textbf{\#Points} & \textbf{\#Edges} & \textbf{Avg./ max aspect ratio} & \textbf{Max height (mm)} & \textbf{Max Displacement} \\
\midrule
\textbf{Mean}  & 20,972 & 114,140 & 1.6421  / 1.6421  & 29.429  & 0.29526    \\
\textbf{Std.}  & 12,476 &  68,308 & 0.43794 / 0.43794 & 23.246  & 0.21064    \\
\textbf{Min}   &    736 &   2,860 & 1.0667  / 1.0667  & 0.60000 & 0.00048500 \\
\textbf{25\%}  & 10,229 &  56,208 & 1.2800  / 1.2800  & 7.8000  & 0.13827    \\
\textbf{50\%}  & 19,743 & 107,680 & 1.4733  / 1.4733  & 21.600  & 0.27075    \\
\textbf{75\%}  & 30,503 & 166,250 & 1.9072  / 1.9072  & 60.000  & 0.41962    \\
\textbf{Max}   & 47,542 & 249,930 & 2.9932  / 2.9933  & 60.000  & 0.99777    \\
\bottomrule
\end{tabular}
\label{tab:lpbf_stats}
\end{table}

\begin{figure}[t]
\centering
\includegraphics[width=0.9\linewidth, trim={0 3pt 0pt 0pt}]{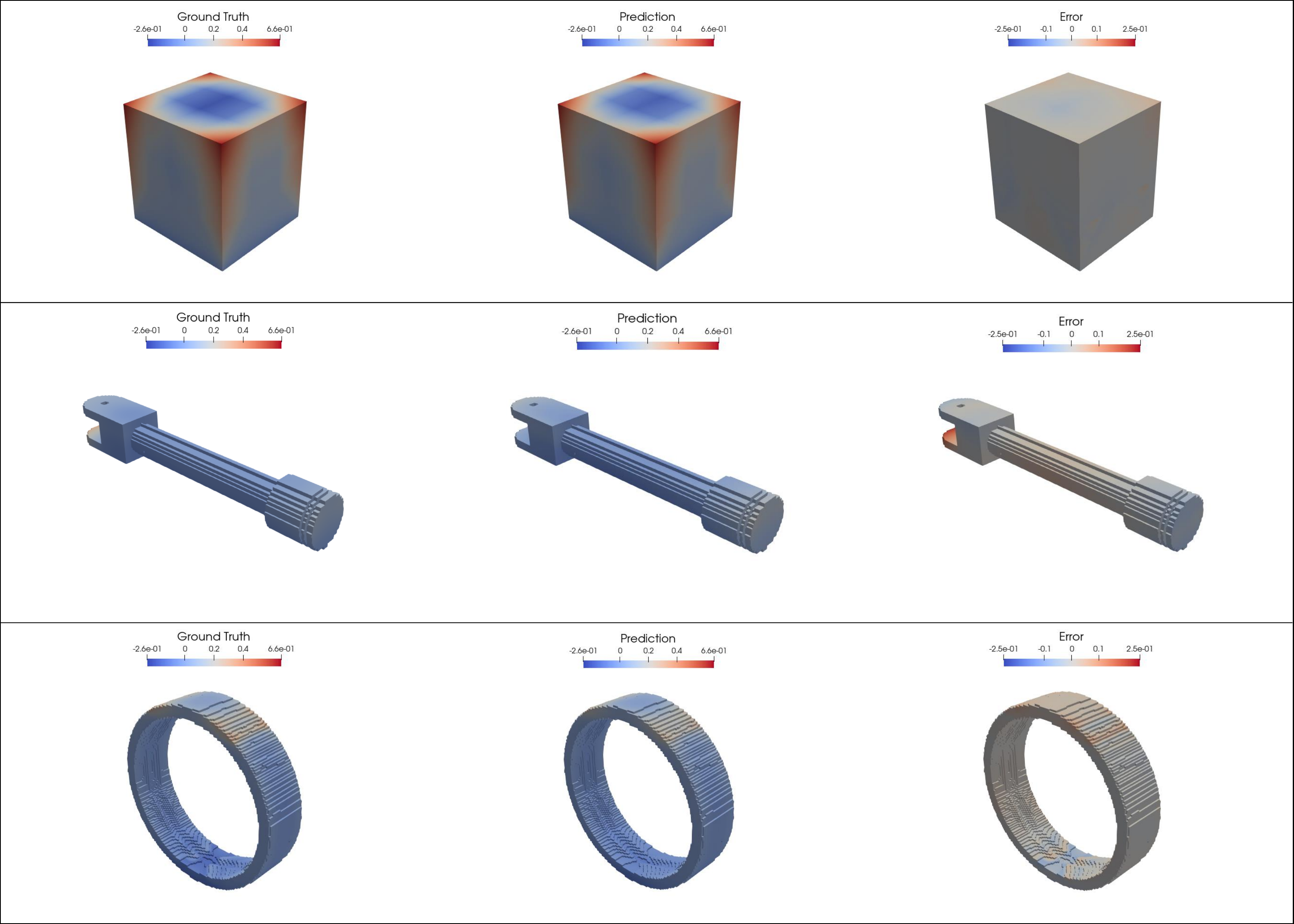}
\includegraphics[width=0.9\linewidth, trim={0 0pt 0pt 3pt}]{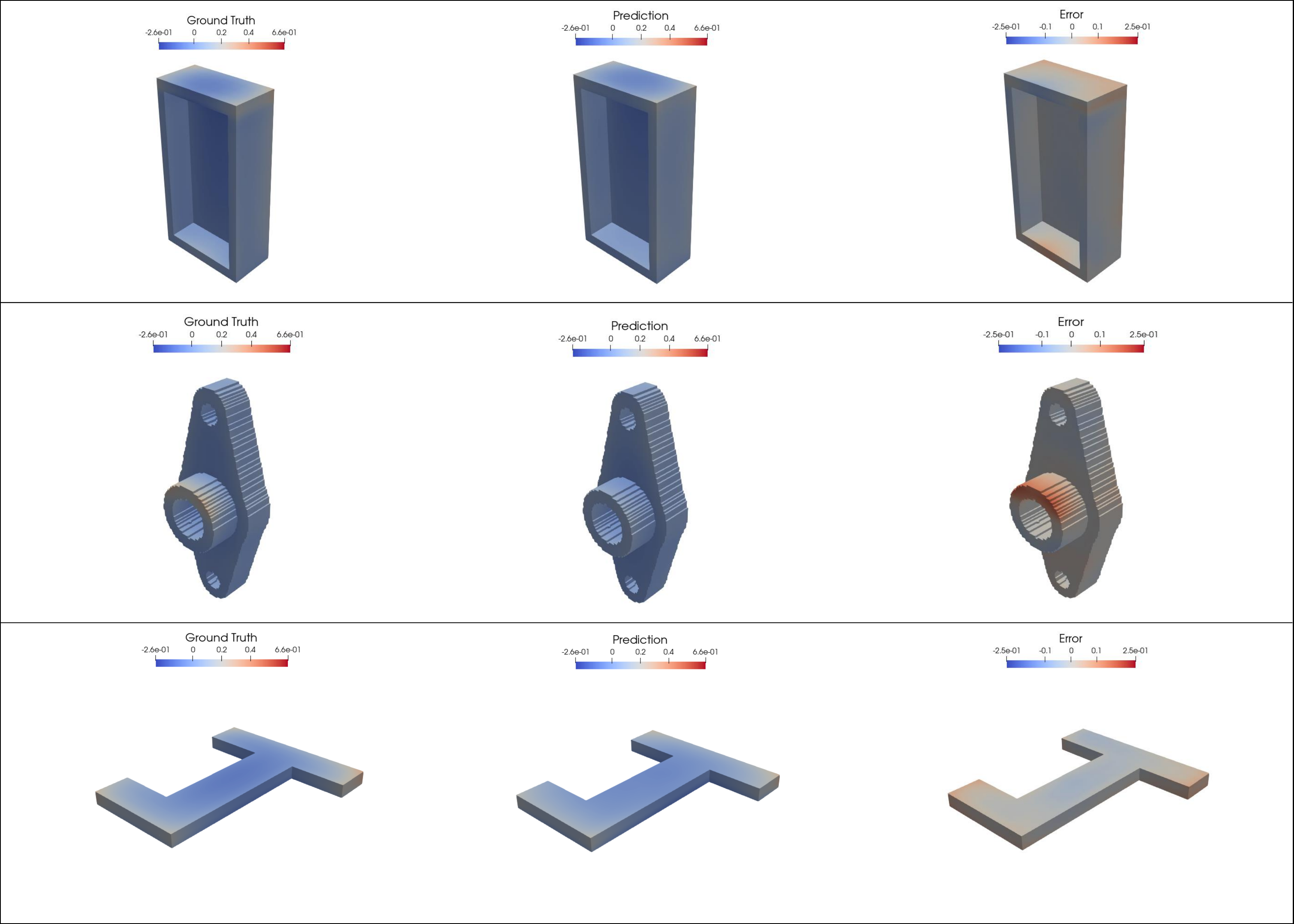}
\caption{
Qualitative results for FLARE on the LPBF $Z$-displacement prediction task.
For each test geometry,
we show the ground truth $Z$-displacement field, the model prediction, and the corresponding error (Ground Truth $-$ Prediction).
}
\label{fig:lpbf_flare_results}
\end{figure}

\subsection{Dataset Generation}

To generate a dataset of LPBF simulations, we employ Autodesk NetFabb \citep{autodesk_help_nodate}, a commercially available software tool for numerically simulating RS and associated physics.
We begin with the Fusion 360 segmentation dataset \citep{lambourne2021brepnet}, and scale each shape to lie within
$[-30, 30] \times [-30, 30] \times [0, 60]~\text{mm}$
such that it rests atop the build plate at $z \in [-25, 0]~\text{mm}$; the parts are not otherwise rotated or transformed.
The simulation is then carried out for the
{Renishaw AM250} machine
and \textit{Ti-6Al-4V} material system deposited with
{$40~\mu\text{m}$} thickness.
Other parameters are left as their default nominal values and no support structures are added \citep{ferguson2025topology}.

In AM, the material is deposited layer by layer, and ideally, a high-fidelity simulation would model each layer individually.
However, this can be computationally expensive, especially for builds with hundreds or thousands of layers.
Layer lumping simplifies this process by combining multiple physical layers into a single \emph{lumped} computational layer.
In our calculations, NetFabb applies layer-lumping with a lumped layer thickness of $2.5~\text{mm}$.

NetFabb re-meshes the geometry to contain axis-aligned hexahedral elements before simulating the build process.
Then, NetFabb generates a thermal and a mechanical history for each part
corresponding to lumped layer deposition steps during the build.
We obtain the displacement, elastic strain, von Mises stress, and temperature fields evaluated at all nodal locations throughout the build.
NetFabb also provides field values after the part has cooled down and detached from the build plate.

\subsection{Benchmark task}
\label{sec:appendix_lpbf_task}

To evaluate neural surrogate models, we target a field prediction task that is both central to our dataset and broadly relevant to the AM community: predicting residual vertical ($Z$) displacement.
In LPBF, each layer is fused by a laser and followed by a recoater blade that spreads powder uniformly across the build area \citep{reijonen2024effect}.
Overhanging features may cause vertical displacements that interfere with the blade’s path, potentially leading to collisions. Predicting the $Z$-displacement field can therefore help identify risk of blade collision failures.
Rapid estimation of displacement prior to a build, thus, is highly desirable for design troubleshooting, as severe distortion can render a part unusable.
Moreover, accurate prediction of nodal displacements can help anticipate build failures \citep{ferguson2025topology}.

Since full-scale LPBF simulations are computationally expensive—taking minutes to hours—a fast surrogate model offers a valuable alternative for accelerating AM design.
Accordingly, we train our models to predict the $Z$-displacement at every node at the final time step.

More formally,
the input to a neural surrogate model is the volumetric axis-aligned hexahedral mesh describing the geometry.
This includes the point-coordinates (array of size $N \times 3$ where $N$ is the number of points) and, optionally, mesh connectivity information.
The corresponding label is the $Z$ displacement value at each point (array of size $N \times 1$).

While this dataset focuses on a steady-state prediction problem, future iterations of this benchmark could involve learning dynamic surrogate models that track the time-history of stress and deformation fields during the build process.

\subsection{Data filtering}

The dataset contains a wide-ranging set of shapes, making the dataset general enough to train a strong data-driven field prediction model.
Out of $\sim$27,000 shapes, 19,732 were successfully simulated.
We analyze the first 3,500 successful simulations and filter them according to several statistics to design a balanced training and test set.
For example, we limit the learning problem to meshes with up to 50,000 points and up to 300,000 edges.
This is done to reduce memory usage which becomes a bottleneck when training on small GPUs.
We also filter meshes that have high aspect ratio elements as the FEM calculation could be unreliable on highly distorted geometries.
Our dataset processing code is available in the GitHub repository associated with this paper.
The statistics for the filtered dataset are presented in \autoref{tab:lpbf_stats} along with histograms in \autoref{fig:lpbf_stats}.
A gallery of successful simulations is presented in \autoref{fig:lpbf_gallery}.


\subsection{Qualitative Results for LPBF $Z$-Displacement Prediction}

In \autoref{fig:lpbf_flare_results}, we present visualizations of ground-truth $Z$-displacement, predictions by FLARE, and the corresponding error.
Across a set of representative geometries from the LPBF test set, FLARE produces displacement fields that are visually indistinguishable from the ground truth and capture both the global deformation patterns and localized high-gradient regions.
Importantly, the error fields remain low-magnitude and spatially diffuse, with no systematic accumulation near edges or corners—indicating that the model does not rely on positional shortcuts or overfit to particular geometric artifacts.
This suggests that FLARE successfully learns the dominant thermo-mechanical modes of deformation while only missing small higher-order variations that are difficult to resolve without substantially larger latent bottlenecks.
These qualitative observations are consistent with the low average relative $L^2$ error and illustrate that FLARE maintains stable accuracy across a diverse distribution of part geometries.

\subsection{Dataset split and benchmark protocol}
\label{sec:appendix_lpbf_split}

\paragraph{Splits.}
From the full set of 19{,}732 successful LPBF simulations, we construct a filtered subset of cases with up to 50{,}000 points per geometry.
We use 1{,}100 cases for training and 290 cases for testing, as summarized in \autoref{tab:dataset_summary}.

\paragraph{Inputs and outputs.}
For each case, the benchmark input consists of the 3D mesh node coordinates $\in\R^{N\times 3}$.
The benchmark target is the final-time-step vertical displacement field ($Z$-displacement) $\in\R^{N\times 1}$ at the same nodes.
We focus on this scalar displacement because it is directly tied to recoater-blade interference risk in LPBF.

\paragraph{Simulation context.}
All parts are scaled to lie within a standardized build volume $[-30, 30] \times [-30, 30] \times [0, 60]~\text{mm}$ and simulated in Autodesk NetFabb for the Renishaw AM250 machine and Ti-6Al-4V material system, using $40~\mu\text{m}$ layer thickness and layer lumping with a lumped layer height of $2.5~\text{mm}$.
These settings follow the procedure described in \autoref{sec:appendix_am_dataset}.


\end{document}